\documentclass[journal]{IEEEtran}
\usepackage{amsfonts}
\usepackage{amsmath,epsfig}
\usepackage{amssymb}
\usepackage{amsbsy}
\usepackage{graphicx}
\usepackage{subfigure}
\usepackage{epstopdf}
\usepackage{algorithm}
\usepackage{algorithmic}
\usepackage{threeparttable}
\usepackage{url}
\usepackage{multirow}
\usepackage{cite}

\begin{document}

\title{DML-GANR: Deep Metric Learning with Generative Adversarial Network Regularization for High Spatial Resolution Remote Sensing Image Retrieval}
\author{}

\author{Yun Cao, Yuebin Wang, $\emph{Member}$, $\emph{IEEE}$, Junhuan Peng, Liqiang Zhang, Linlin Xu, Kai Yan and Lihua Li%
\thanks{Y. Cao, J. Peng, L. Xu, K. Yan and L. Li are with the School of Land Science and Technology, China
University of Geosciences, Beijing 100083, China, and also with the Shanxi Provincial Key Laboratory
of Resources, Environment and Disaster Monitoring, Jinzhong 030600, China (e-mail: cy12160019@163.com;  pengjunhuan@163.com; beyond13031@126.com; kyan@mail.bnu.edu.cn; lihuali@cugb.edu.cn).}
\thanks{Y. Wang is with the School of Land Science and Technology, China University of Geosciences, Beijing 100083, China, and also with State Key Laboratory of Remote Sensing Science, Beijing Normal University, Beijing 100875, China (e-mail: xxgcdxwyb@163.com).}
\thanks{L. Zhang is with the Faculty of Geographical Science, State Key Laboratory of Remote Sensing
Science, Beijing Normal University, Beijing 100875, China (e-mail: zhanglq@bnu.edu.cn).}}


\maketitle

\begin{abstract}
\boldmath
With a small number of labeled samples for training, it can save considerable manpower and material resources, especially when the amount of high spatial resolution remote sensing images (HSR-RSIs) increases considerably. However, many deep models face the problem of overfitting when using a small number of labeled samples. This might degrade HSR-RSI retrieval accuracy. Aiming at obtaining more accurate HSR-RSI retrieval performance with small training samples, we develop a deep metric learning approach with generative adversarial network regularization (DML-GANR) for HSR-RSI retrieval. The DML-GANR starts from a high-level feature extraction (HFE) to extract high-level features, which includes convolutional layers and fully connected (FC) layers. Each of the FC layers is constructed by deep metric learning (DML) to maximize the interclass variations and minimize the intraclass variations. The generative adversarial network (GAN) is adopted to mitigate the overfitting problem and validate the qualities of extracted high-level features. DML-GANR is optimized through a customized approach and the optimal parameters are obtained. Experimental results on three data sets demonstrate the superior performance of DML-GANR over state-of-the-art techniques in HSR-RSI retrieval.
\end{abstract}


\section{Introduction}
\label{sec:introduction}

\IEEEPARstart{W}{ith} the accessibility of large volumes of HSR-RSIs acquired by satellites and airborne sensors, HSR-RSI processing and analysis have been an active topic in recent years \cite{justice1998the,Yu2006Object,thomas2008synthesis,blaschke2010object,hu2015transferring}. The image retrieval task is one of the most challenging and basic technologies \cite{ferecatu2007interactive,aptoula2014remote,demir2015a,li2018large-scale}. Due to a large number of images, it is important to develop effective and efficient methods to search, retrieve and recognize images that are similar to the query image. It is known that using inappropriate methods for feature descriptor quantization might lead to a significant degradation in image retrieval \cite{rubner2000the}.

The HSR-RSI retrieval suffers from intraclass diversity and interclass similarity that often degenerate the HSR-RSI retrieval performance. In this situation, it is important to learn representative and discriminative feature representations that have small intraclass scatter but large interclass separation. Metric learning can offer a solution for image retrieval, where metric distances provide a measure of the dissimilarity among different data points \cite{weinberger2009distance}. Supervised distance metric learning algorithms learn a distance metric that can minimize the variations of samples from the same class and maximize the separability of samples from different classes \cite{wang2017learning}. The image retrieval accuracy can be improved significantly with appropriately designed distance metrics. The image-to-class distance metric learning method for image classification was suggested in \cite{wang2010image}, and the discrimination of the image-to-class distance was improved by learning the per-class Mahalanobis metrics. Cheng $\emph{et al.}$ \cite{cheng2018exploring} proposed a unified metric learning-based framework to learn more representative features for hyperspectral image classification. Wang $\emph{et al.}$ \cite{wang2017learning} proposed a discriminative distance metric learning method with label consistency for HSR-RSI scene classification, where dense scale invariant feature transformation and spatial pyramid maximum pooling with sparse coding were used to extract and encode features from HSR-RSIs, respectively. From the above tasks, metric learning plays an important role in improving image retrieval accuracy and scene classification accuracy.

However, many traditional metric learning algorithms usually use a single linear distance to transform samples into a linear feature space, which cannot exploit the deep and nonlinear relationship of samples well. Deep learning feature learning methods extract features in a deep way that can obtain powerful and discriminative high-level feature representations than low-level features from handcrafted feature descriptors. More recently, metric learning has been combined with related deep learning methods. In \cite{cai2012deep}, the nonlinear metric learning method was proposed with stacked independent subspace analysis. A discriminative DML was proposed by Hu $\emph{et al.}$ \cite{hu2014discriminative}, which trains a deep neural network by learning nonlinear transformations. Under the discriminative DML, the face pairs were projected into other feature subspaces. The distance of each positive face pair is less than a smaller threshold, while the distance of each negative pair is higher than a larger threshold. The key advantage of DML is that the deep neural network is used to learn the nonlinear distance metric \cite{hu2014discriminative}. The nonlinear mappings are explicitly obtained, and a back-propagation algorithm can be used to train the network.

To obtain accurate HSR-RSI retrieval performance, many labeled scene images are usually needed. However, the manual generation of tags is usually time-consuming, labor intensive, expensive, and sometimes causes error tags \cite{li2018large-scale}. Such a task becomes harder to execute when the number of HSR-RSIs increases considerably. One solution is to use a small number of labeled samples for training. However, due to the imbalance between millions of parameters and limited training samples, the application of deep CNNs for tasks is prone to overfitting \cite{ioffe2015batch,wang2016stct,cheng2017duplex,cheng2018learning}. Adding regularization is one of the general tools for reducing overfitting. In \cite{wang2016stct}, each channel of the output feature map is trained with different loss criteria to avoid overtraining. Cheng $\emph{et al.}$ \cite{cheng2017duplex} proposed a metric learning regularization term to force the features to have small intraclass scatter and large interclass separation. In \cite{cheng2018learning}, rotation-invariant regularization and Fisher discrimination regularization were imposed on the CNN features.

As indicated by recent works, GAN regularization \cite{goodfellow2014generative} can be used to mitigate the overfitting problem. GAN consists of two models: a generator model $\emph{G}$ and a discriminative model $\emph{D}$. The input of the generator $\emph{G}$ can be any dimensional tensor, while the output of the generator $\emph{G}$ is similar to the real dataset. The generated images, which are also called ``adversarial samples'', have the same label as the corresponding real images. As such, the generated images increase the number of labeled training images and enrich the diversity of the training samples. In addition, when the HSR-RSI retrieval accuracy is limited with small training samples, GAN regularization can be adopted to determine the quality of extracted high-level features.

\begin{figure*}[htb]
\centering\includegraphics[width=7.1 in]{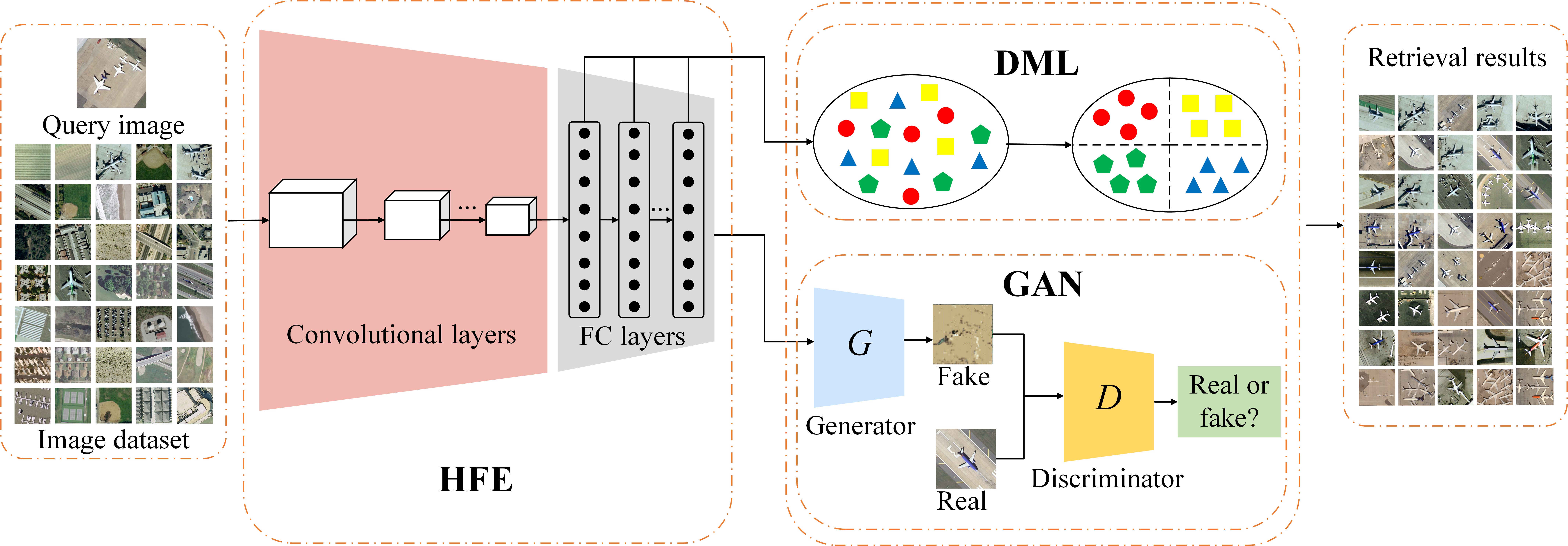}
\caption{Architecture of proposed DML-GANR.}\label{overallDML-GANR}
\end{figure*}

Inspired by the above discussions, we develop a novel feature extraction network DML-GANR for HSR-RSI retrieval. Multilayer DML is introduced to mitigate the intraclass diversity and interclass similarity problem. GAN regularization is further introduced to mitigate the overfitting problem caused by the small training datasets. Moreover, GAN can evaluate the qualities of the extracted features using the generated image. In this paper, we first employ HFE to extract basic image features, which consists of two parts: the convolutional layers and the FC layers. Then, the basic features obtained by HFE are provided for DML. The multilayer DML is used to increase the intraclass compactness and interclass separability for HSR-RSI retrieval. Moreover, the extracted features are further processed with GAN. In the GAN, the generator $\emph{G}$ accepts the extracted features as inputs and synthesizes fake HSR-RSIs that are similar to real HSR-RSIs. The discriminator $\emph{D}$ accepts either real data or fake data as inputs and discriminates where the inputs came from. The DML-GANR is optimized with a customized approach to obtain representative features. In the experiment, our method is compared with the related approaches. This indicates that our method can achieve better retrieval results than other methods. An overview of DML-GANR for HSR-RSI retrieval is shown in Fig. \ref{overallDML-GANR}.

The main contributions of our paper are summarized as follows:

1) A novel deep learning approach for HSR-RSI retrieval called DML-GANR is developed, in which part of the multilayer DML and GAN are integrated into a unified objective function.

2) To mitigate the overfitting problem of small samples on DML and further improve the accuracy of HSR-RSI retrieval, we introduce GAN. The adversarial samples generated by generator $\emph{G}$ not only increase the number of labeled training images and enrich the diversity of the training samples but can also  be used to evaluate the qualities of the extracted features.

3) DML-GANR is optimized through a customized approach and obtained the optimal parameters. Experimental results on the three datasets demonstrate the superior performance of the DML-GANR over state-of-the-art techniques, especially on small samples.

\section{Related Work}
\label{sec:Related Work}

In this section, the related work about the deep CNNs, GAN, metric learning, and HSR-RSI retrieval is introduced.
\subsection{Deep Convolutional Neural Networks (CNNs)}
\label{sec:Deep Convolutional Neural Networks (CNNs)}

Deep CNNs have proven to be effective for large-scale visual recognition on ImageNet, such as AlexNet \cite{krizhevsky2012imagenet}, VGGNet \cite{simonyan2014very}, GoogLeNet \cite{szegedy2014going}, and ResNet \cite{he2016deep}, which can extract mid- and high- level features from raw images by spatially shrinking the feature maps layer by layer \cite{Zhu2017Deep}. Recently, a few works have investigated pretrained CNNs on ImageNet that can be applied to HSR-RSI feature extraction. Castelluccio $\emph{et al.}$ \cite{castelluccio2015land} used a pretrained network and fine-tuned the training data for the HSR-RSI land use classification. Hu $\emph{et al.}$ \cite{hu2015transferring} proposed two scenarios for scene classification. One scenario selects the activation vectors extracted from the FC layers as the final image features. The other encodes dense features extracted from the last convolutional layer at multiple scales into global image features. Marmanis $\emph{et al.}$ \cite{marmanis2016deep} used the pre-trained CNN (i.e., Overfeat model \cite{sermanet2013overfeat:}) to extract the initial set of representations that are then transferred into a supervised CNN classifier. This method can effectively address large data dimensionality. Han $\emph{et al.}$ \cite{han2017pre-trained} proposed an improved pretrained AlexNet framework that combines scale pooling spatial pyramid pooling and side supervision to improve the classification results and better deal with the multiscale information of the convolved feature maps of HSR-RSI. In \cite{han2017a}, the deep learning feature, a cotraining-based self-label technique, and a discriminative evaluation are combined for HSR-RSI semisupervised scene classification. The two independent pretrained CNNs (i.e., ResNet-50 and VGG-S) are used to generate high-level feature representations and compared with CaffeNet, GoogLeNet, VGG, and ResNet-50. In \cite{nogueira2017towards}, three strategies of exploiting existing CNNs (i.e., fully-trained CNNs, fine-tuned CNNs, and pretrained CNNs) were evaluated and analyzed. The results showed that fine-tuning outperforms competing strategies.

\subsection{Generative Adversarial Network (GAN)}
\label{sec:Generative Adversarial Network (GAN)}

GANs are a new powerful class of networks introduced by Goodfellow $\emph{et al.}$ \cite{goodfellow2014generative}, which include a generator $\emph{G}$ and a discriminator $\emph{D}$. The generator $\emph{G}$ learns the data probability distribution and generates new fake samples from that distribution. The discriminator $\emph{D}$ determines whether the input sample comes from fake data or real data. The generator $\emph{G}$ and the discriminator $\emph{D}$ are trained simultaneously, and the purpose of GAN optimization is to reach a Nash equilibrium \cite{ratliff2013characterization}, where $\emph{G}$ is capable of creating the real data probability and $\emph{D}$ cannot distinguish between fake data and real data.

There are a few studies of GAN in image retrieval. Song $\emph{et al.}$ \cite{song2018binary} used a binary GAN for image retrieval that can simultaneously learn a binary representation per image and generate an image similar to the real data. A new sign activation strategy and a loss function are designed, and the loss function consists of an adversarial loss, a content loss, and a neighborhood structure loss. The use of GAN in remote sensing has proven to be successful. Lin $\emph{et al.}$ \cite{lin2017marta} proposed an unsupervised multilayer feature-matching GAN to learn a feature representation using only unlabeled data for the HSR-RSI classification. The generator $\emph{G}$ produces numerous images that are similar to the real data, and the discriminator $\emph{D}$ extracts features. A fusion layer is used to merge the midlevel and global features for fitting the complex properties of the HSR-RSI. Shi $\emph{et al.}$ \cite{shi2018road} proposed a novel end-to-end GAN network for road detection from the HSR-RSI, which obtains the segmentation results by finding and correcting the difference between the ground truth and the generated result by the generator $\emph{G}$.

\subsection{Metric Learning}
\label{sec:Metric Learning}

A similarity measure is a function that defines the distance between the extracted features to identify images similar to the query \cite{aptoula2014remote}. Distance metric learning is an ideal alternative for manually constructing a similarity metric, that is capable of automatically learning distance functions and has proven to be useful in image retrieval \cite{hoi2010semi,lee2008rank,huang2012large,si2006collaborative}. The purpose of distance metric learning is to maximize the interclass variation and minimize the intraclass variation \cite{zantedeschi2016metric}. Hoi $\emph{et al.}$ \cite{hoi2010semi} proposed a semisupervised distance metric learning framework named Laplacian regularized metric learning for collaborative image retrieval and clustering. The unlabeled data are incorporated into the distance metric learning task through a regularized learning framework. Huang $\emph{et al.}$ \cite{huang2012large} addressed the problems of learning Mahanalobis distance metrics in a high-dimensional feature space by proposing an ensemble metric learning method. The proposed method includes sparse block diagonal metric ensembling and joint metric learning for face verification and retrieval. A new deep transfer metric learning method was proposed by Hu $\emph{et al.}$ \cite{hu2015deep} to learn a set of nonlinear transformations that transfer discriminative knowledge from the labeled source domain to the unlabeled target domain. Han $\emph{et al.}$ \cite{han2017unified} proposed a unified metric learning-based framework to learn more representative representations, which embeds metric learning regularization into the support vector machine. When metric learning is applied in HSR-RSI domains, Wang $\emph{et al.}$ \cite{wang2017learning} proposed a discriminative distance metric learning method with label consistency for HSR-RSI scene classification, which optimizes the method by using the joint optimization of feature manifold, distance metric, and label distribution. In \cite{gong2018diversity-promoting}, diversity-promoting deep structural metric learning was incorporated into deep networks through a structured loss for HSR-RSI scene classification. In \cite{cheng2018when}, a metric learning regularization term imposed on the CNN features for the HSR-RSI scene classification was trained by optimizing a new discriminative objective function.

\subsection{HSR-RSI Retrieval}
\label{sec:HSR-RSI retrieval}

The retrieval performance of the HSR-RSI mostly depends on the extracted features. Recently, deep learning methods have demonstrated their excellent capacity to extract powerful feature representations \cite{zhou2018patternnet:}. In \cite{cheriyadat2014unsupervised}, an unsupervised feature learning method with SIFT and sparse coding was proposed to generate a set of functions from unlabeled features for aerial scene classification. Zhou $\emph{et al.}$ \cite{zhou2015high-resolution} proposed an unsupervised sparse feature learning framework that uses an autoencoder network for HSR-RSI retrieval. Wang $\emph{et al.}$ \cite{wang2016a} proposed a novel graph-based learning method based on a three-layer framework for retrieving HSR-RSI, which integrates the strengths of query expansion and the fusion of holistic and local features. This framework effectively uses the potential tag information of the HSR-RSI, which achieves predominant performance on experimental datasets. In \cite{li2016content-based}, a multiple feature representation and collaborative affinity metric fusion approach was proposed for HSR-RSI retrieval. In this approach, four unsupervised CNNs were designed to generate four types of unsupervised features from the fine level to the coarse level. Penatti $\emph{et al.}$ \cite{penatti2015deep} investigated the generalization power of deep features extracted by CNNs, which transfers deep features from everyday objects to remote sensing. Napoletano \cite{napoletano2018visual} extensively evaluated visual descriptors for content-based HSR-RSI retrieval, which includes global handcrafted, local hand-crafted, and CNN features coupled with four different content-cased image retrieval schemes. Experimental results show that CNN-based features outperform both global and local hand-crafted features. Zhou $\emph{et al.}$ \cite{zhou2017learning} studied how to extract powerful feature representations based on the pretrained CNN for HSR-RSI retrieval. There are two schemes: one scheme extracts deep features from the fully connected and convolutional layers of the pretrained CNN models, and the other is a novel CNN architecture based on conventional layers and a three-layer perceptron.

\section{Deep Metric Learning with Generative Adversarial Network Regularization (DML-GANR)}
\label{sec:Deep Metric Learning with Generative Adversarial Network Regularization (DML-GANR)}

In this section, we first present the motivations of the proposed DML-GANR and subsequently present the procedure and the optimization of the DML-GANR.

\subsection{Problem Formulation}
\label{sec:Problem Formulation}

To obtain accurate HSR-RSI retrieval performance in the case of small training samples, the DML-GANR is proposed in this paper. The challenge of small training samples is the overfitting problem. GAN can mitigate the overfitting problem by generating adversarial samples. The adversarial samples generated by the generator $\emph{G}$ not only increase the number of labeled training images and enrich the diversity of the training samples but can also be used to evaluate the qualities of the extracted features. There are three parts in DML-GANR: HFE, multilayer DML, and GAN. Features extracted from HFE are provided for DML to learn a metric in a supervised way, which can measure
the similarity among the training dataset. Moreover, the extracted features are further processed with the GAN to generate adversarial samples.

To improve the clarity of this paper, we illustrate important notations and definitions in Table \ref{NOTATIONS AND DEFINITIONS}.

\renewcommand\arraystretch{1.25}
\begin{table}[htbp]
\caption{Notations and Definitions }
\centering{}
\begin{tabular}{c c}
\hline
Notation & Definitions \\
\hline
$\textbf{X}$ & Raw images.\\
\hline
$\textbf{W}_{\emph{F}}^{(\emph{l})}$ & The weight matrix of FC layers. 1$\leqslant \emph{l} \leqslant \emph{L}$.\\
\hline
$\textbf{b}_{\emph{F}}^{(\emph{l})}$ & The basis of FC layers. 1$\leqslant \emph{l} \leqslant \emph{L}$.\\
\hline
$\textbf{u}^{(\emph{l})}$ & The extracted high-level features. 1$\leqslant \emph{l} \leqslant \emph{L}$.\\
\hline
$\emph{F}\left ( \cdot  \right )$ & The convolutional layers of HFE. \\
\hline
$\left \| \cdot  \right \|_\emph{F}$ & The Frobenius norm. \\
\hline
$\emph{S}_{\emph{c}}^{(\emph{l})}$ & The intra-class compactness. 1$\leqslant \emph{l} \leqslant \emph{L}$.\\
\hline
$\emph{S}_{\emph{b}}^{(\emph{l})}$ & The inter-class separability. 1$\leqslant \emph{l} \leqslant \emph{L}$.\\
\hline
$\emph{G}$ & The generator of GAN.\\
\hline
$\emph{D}$ & The discriminator of GAN.\\
\hline
$\textbf{H}_{\emph{G}}^{(\emph{h})}$ & The output of $\emph{h}$ layer of $\emph{G}$.\\
\hline
$\textbf{H}_{\emph{D}}^{(\emph{k})}$ & The output of $\emph{k}$ layer of $\emph{D}$.\\
\hline
\multirow{2}{*}{$\textbf{W}_{\emph{G}}^{(\emph{h})}$, $\textbf{W}_{\emph{D}}^{(\emph{k})}$} & The weight matrix of $\emph{G}$ and $\emph{D}$, respectively. \\
&1$\leqslant \emph{h} \leqslant \emph{H}$, 1$\leqslant \emph{k} \leqslant \emph{K}$.\\
\hline
\multirow{2}{*}{$\textbf{b}_{\emph{G}}^{(\emph{h})}$, $\textbf{b}_{\emph{D}}^{(\emph{k})}$} & The basis of $\emph{G}$ and $\emph{D}$, respectively. \\
&1$\leqslant \emph{h} \leqslant \emph{H}$, 1$\leqslant \emph{k} \leqslant \emph{K}$.\\
\hline
$\lambda $, $\alpha $ & Balance the corresponding terms.\\
\hline
$\gamma $ & A tunable positive regularization parameter.\\
\hline
$\delta $ & The learning rate of DML.\\
\hline
$\beta _{1}$, $\beta _{2}$ & The learning rate of GAN.\\
\hline
$\varphi(\cdot )$, $\psi(\cdot )$, $\text{Tanh}(\cdot )$, $\sigma (\cdot )$ & The activation functions.\\
\hline
$\emph{P}(\cdot ) $, $\text{up}(\cdot )$  & The max-pooling and its reversion.\\
\hline
$\emph{R}(\cdot ) $ &  The rotation of the input by 180 degree.\\
\hline
\end{tabular}
\label{NOTATIONS AND DEFINITIONS}
\end{table}

\subsection{DML-GANR}
\label{sec:DML-GANR}

\subsubsection{High-level Feature Extraction (HFE)}
\label{sec:High-Level Feature Extraction (HFE)}

High-level features are introduced to describe HSR-RSIs, which are extracted by our feature extraction model HFE. There are two parts in HFE: the convolutional layers with a fixed set of weights and no labels and the FC layers with an unfixed set of weights and no labels. In this section, we present the procedure for HFE.

The first component of the HFE model is the convolutional layers; specifically, a pretrained CNN model. In detail, we employ ResNet-50 \cite{he2016deep} to extract the high-level features from HSR-RSIs, which is proven to achieve the best performance among various CNNs (i.e., AlexNet, CaffeRef, VGG, VGG-VD, and GoogLeNet) \cite{zhou2018patternnet:}. In the ResNet framework \cite{he2016deep}, the authors present a residual learning framework to avoid the problem of vanishing gradients. For a mathematical expression, input image $\textbf{X}$ is transformed by convolutional layers $\emph{F}({\textbf{X}})$, and then $\textbf{X}$ is added to $\emph{F}({\textbf{X}})$. After the activation function such as ReLU $\varphi \left ( \cdot  \right )$, the output $\textbf{y}$ equals:
\begin{equation}
\begin{aligned}
\textbf{y} = \varphi \left ( \textbf{X} + \emph{F}({\textbf{X}}) \right )
\label{Resnet}
\end{aligned}
\end{equation}

The training HSR-RSI are first fed into ResNet-50, and then features $\textbf{u}^{(0)}$ are extracted from the 50th FC layer of ResNet-50. In the training process, the parameters of earlier layers of ResNet-50 are fixed, and the parameters of the 50th FC layer are adjusted to improve the accuracy.

The second component of HFE is multiple FC layers. The importance of the FC layers is to improve access to more representative high-level features. The FC layers accept the previously derived features $\textbf{u}^{(0)}$ as an input. We feed features $\textbf{u}^{(0)}$ into multiple FC layers to obtain multiple sets of high-level features.
\begin{equation}
\begin{aligned}
\textbf{u}^{(\emph{l})} = \psi \left ( \textbf{W}_{\emph{F}}^{(\emph{l})}\cdot \textbf{u}^{(\emph{l-}1)} + \textbf{b}_{\emph{F}}^{(\emph{l})} \right )
\label{FC}
\end{aligned}
\end{equation}

\noindent where $\emph{l}$ denotes the $\emph{l}$th layer (1$\leqslant$ $\emph{l}$ $\leqslant$ $\emph{L}$). $\textbf{W}_{\emph{F}}^{(\emph{l})}$ (1$\leqslant \emph{l} \leqslant \emph{L}$) is the weight matrix of the $\emph{l}$th layer, and $\textbf{b}_{\emph{F}}^{(\emph{l})}$ (1$\leqslant \emph{l} \leqslant \emph{L}$) is the bias vector of the $\emph{l}$th layer. $\cdot $ represents FC multiplication, and $\psi (\cdot )$ is the LeakyReLU (LReLU) activation function. For the first layer, we assume the $\textbf{u}^{(\emph{l-}1)} = \textbf{u}^{(0)}$.

These features extracted from multiple FC layers are assigned to high-level features. The features $\textbf{u}^{(\emph{l})}$ (1$\leqslant \emph{l} \leqslant \emph{L}$) are input into the DML to calculate the distances among the features of the training images. In addition, the features $\textbf{u}^{(\emph{L})}$ are input into GAN to evaluate the quantity of extracted features.

\subsubsection{Multilayer DML}
\label{sec:Multi-Layer DML}

Since the high-level features $\textbf{u}^{(\emph{l})}$ have been obtained, the multilayer DML learns a distance metric among the features of the training images to make dissimilar images far from each other and similar images close to each other. As introduced in Section \ref{sec:High-Level Feature Extraction (HFE)}, the high-level features are extracted through multiple FC layers of HFE. Each FC layer is constructed by DML. For each pair of samples $\textbf{X}_{\emph{i}}$ and $\textbf{X}_{\emph{j}}$, they can be finally represented as $\textbf{u}_{\emph{i}}^{(\emph{l})}$ and $\textbf{u}_{\emph{j}}^{(\emph{l})}$ at the $\emph{l}$th layer (1$\leqslant \emph{l} \leqslant \emph{L}$). Given the features $\textbf{u}_{\emph{i}}^{(\emph{l})}$ and $\textbf{u}_{\emph{j}}^{(\emph{l})}$, we measure their distance metric by computing the squared Euclidean distance between $\textbf{u}_{\emph{i}}^{(\emph{l})}$ and $\textbf{u}_{\emph{j}}^{(\emph{l})}$ at the $\emph{l}$th layer:
\begin{equation}
\begin{aligned}
\emph{d}^2_{\textbf{u}^{(\emph{l})}}\left (  \textbf{u}_{\emph{i}}^{(\emph{l})},\textbf{u}_{\emph{j}}^{(\emph{l})}\right ) = \left \|  \textbf{u}_{\emph{i}}^{(\emph{l})}- \textbf{u}_{\emph{j}}^{(\emph{l})} \right \|^2_2
\label{DMLdistance}
\end{aligned}
\end{equation}

The categories of some scene images are often obscured by each other. Traditional distance metrics cannot distinguish these images clearly. Finding a suitable distance metric can enlarge the distances among dissimilar images and make similar images more compact. Thus, a new distance metric is needed to learn the similarity between the query image and the retrieved images. To achieve this, we employ the margin constraint to learn the metric. The margin constraint includes two terms: the first term is to achieve intraclass compactness, and the second term is to control interclass separability. For each feature extracted from the HFE, we calculate and sum the corresponding margin constraints among the features of the training images and then formulate a supervised multilayer DML that can be described as follows:
\begin{equation}
\begin{aligned}
\Phi _{\text{DML}} = \sum_{l=1}^{L} \left [ \emph{S}_{\emph{c}}^{(\emph{l})} - \alpha  \emph{S}_{\emph{b}}^{(\emph{l})} + \gamma \left (  \left \| \textbf{W}_{\emph{F}}^{(\emph{l})} \right \|_{\emph{F}}^2 + \left \| \textbf{b}_{\emph{F}}^{(\emph{l})} \right \|_{\emph{F}}^2 \right )
\right ]
\label{lossDML}
\end{aligned}
\end{equation}

\noindent where 1$\leqslant \emph{l} \leqslant \emph{L}$. $\alpha $ ($\alpha > $ 0) is the parameter to balance the contributions of the first term and the second term, and $\gamma$ is a tunable positive regularization parameter. $\left \| \cdot  \right \|_{\emph{F}}$ is the Frobenius norm. $\emph{S}_{\emph{c}}^{(\emph{l})}$ defines the intraclass compactness, and $\emph{S}_{\emph{b}}^{(\emph{l})}$ defines the interclass separability.
\begin{equation}
\begin{aligned}
\emph{S}_{\emph{c}}^{(\emph{l})} = \frac{1}{\emph{n}\emph{t}_1} \sum_{\emph{i}=1}^{\emph{n}} \sum_{\emph{j}=1}^{{\emph{t}}_1} \emph{P}_\emph{ij} \left \|  \textbf{u}_{\emph{i}}^{(\emph{l})}- \textbf{u}_{\emph{j}}^{(\emph{l})} \right \|^2_2
\end{aligned}
\end{equation}

\begin{equation}
\begin{aligned}
\emph{S}_{\emph{b}}^{(\emph{l})} = \frac{1}{\emph{n}\emph{t}_2} \sum_{\emph{i}=1}^{\emph{n}} \sum_{\emph{j}=1}^{{\emph{t}}_2} \emph{Q}_\emph{ij} \left \|  \textbf{u}_{\emph{i}}^{(\emph{l})}- \textbf{u}_{\emph{j}}^{(\emph{l})} \right \|^2_2
\end{aligned}
\end{equation}

\noindent where $\emph{n}$ is the number of labeled training dataset, $\emph{t}_1$ is the number of $\emph{t}_1$-intraclass nearest neighbors of the feature matrix of the query image $\textbf{X}_\emph{i}$, and $\emph{t}_2$ is the number of $\emph{t}_2$-interclass nearest neighbors of the feature matrix of the query image $\textbf{X}_\emph{i}$. $\emph{P}_\emph{ij}$ is set as 1 if $\textbf{X}_\emph{j}$ is one of $\emph{t}_1$-intraclass nearest neighbors of $\textbf{X}_\emph{i}$, and 0 otherwise. $\emph{Q}_\emph{ij}$ is set as 1 if $\textbf{X}_\emph{j}$ is one of $\emph{t}_2$-interclass nearest neighbors of $\textbf{X}_\emph{i}$, and 0 otherwise.

\subsubsection{GAN Regularization}
\label{sec:GAN Regularization}

GAN is introduced to mitigate the overfitting problem and check the quality of high-level features extracted from the HFE. The $\textbf{u}^{(\emph{L})}$ are sent into GAN as the input of the generator $\emph{G}$. The generator $\emph{G}$ builds a mapping from the features $\textbf{u}^{(\emph{L})}$ to the outputting image, which is represented by $\emph{G}(\textbf{u}^{(\emph{L})})$. There are $\emph{H}$ hidden layers in $\emph{G}$, in which the first three layers are FC layers and the subsequent layers are convolutional layers. The output $\textbf{H}_{\emph{G}}^{(\emph{h})}$ at $\emph{h}$th layer (1$\leqslant \emph{h}\leqslant \emph{H}$) is computed as follows:
\begin{equation}
\begin{aligned}
\textbf{H}_{\emph{G}}^{(\emph{h})} = \left \{ \begin{matrix} \varphi \left ( \textbf{W}_{\emph{G}}^{(\emph{h})}\cdot \textbf{H}^{(\emph{h-}1)}_{\emph{G}} + \textbf{b}_{\emph{G}}^{(\emph{h})} \right ), 1\leqslant \emph{h} \leqslant 3
\\ \varphi \left ( \textbf{W}_{\emph{G}}^{(\emph{h})}\otimes  \textbf{H}^{(\emph{h-}1)}_{\emph{G}} + \textbf{b}_{\emph{G}}^{(\emph{h})} \right ), 3 < \emph{h} < \emph{H}
\\ \text{Tanh}\left ( \textbf{W}_{\emph{G}}^{(\emph{h})}\otimes  \textbf{H}^{(\emph{h-}1)}_{\emph{G}} + \textbf{b}_{\emph{G}}^{(\emph{h})} \right ), \emph{h} = \emph{H}
\end{matrix} \right.
\label{HFh}
\end{aligned}
\end{equation}

\noindent where $\textbf{W}_{\emph{G}}^{(\emph{h})}$ is the weight of $\emph{G}$, and $\textbf{b}_{\emph{G}}^{(\emph{h})}$ is the bias of $\emph{G}$. $\varphi \left ( \cdot \right )$ is the ReLU activation function, $\otimes$ represents the convolution operation, and $\text{Tanh}\left ( \cdot \right )$ represents the tanh activation function.

The discriminator $\emph{D}$ gives a probability of whether the input comes from the real data or fake data. Let $\emph{D}(\textbf{X})$ be the probability of the real data, and $\emph{D}(\emph{G}(\textbf{u}^{(\emph{L})}))$ be the probability of the fake data. There are $\emph{K}$ hidden layers in $\emph{D}$, which includes six convolutional layers, two max-pooling layers, one concatenation layer, and several FC layers. The output $\textbf{H}_{\emph{D}}^{(\emph{k})}$ at $\emph{k}$th layer (1$\leqslant \emph{k} \leqslant \emph{K}$) is computed as follows:
\begin{equation}
\begin{aligned}
\textbf{H}_{\emph{D}}^{(\emph{k})} = \left \{ \begin{matrix} &\psi \left ( \textbf{W}_{\emph{D}}^{(\emph{k})} \otimes  \textbf{H}^{(\emph{k-}1)}_{\emph{D}} + \textbf{b}_{\emph{D}}^{(\emph{k})} \right ), 1 \leqslant \emph{k} \leqslant 6
\\ &\emph{P} \left ( \textbf{H}^{(4)}_{\emph{D}} \right ), \emph{k} = 7
\\ &\emph{P} \left ( \textbf{H}^{(5)}_{\emph{D}} \right ), \emph{k} = 8
\\ &\textbf{H}^{(6)}_{\emph{D}} \oplus \textbf{H}^{(7)}_{\emph{D}} \oplus \textbf{H}^{(8)}_{\emph{D}}, \emph{k} = 9
\\ &\psi \left ( \textbf{W}_{\emph{D}}^{(\emph{k})}\cdot \textbf{H}^{(\emph{k-}1)}_{\emph{D}} + \textbf{b}_{\emph{D}}^{(\emph{k})} \right ), 9 < \emph{k} < \emph{K}
\\ &\sigma \left ( \textbf{W}_{\emph{D}}^{(\emph{k})} \cdot \textbf{H}^{(\emph{k-}1)}_{\emph{D}} + \textbf{b}_{\emph{D}}^{(\emph{k})} \right ), \emph{k} = \emph{K}
\end{matrix} \right.
\label{HDk}
\end{aligned}
\end{equation}

\noindent where $\textbf{W}_{\emph{D}}^{(\emph{k})}$ is the weight of $\emph{D}$, and $\textbf{b}_{\emph{D}}^{(\emph{k})}$ is the bias of $\emph{D}$. $\emph{P}(\cdot)$ is the max-pooling function, $\oplus $ is the concatenation symbol, and $\sigma (\cdot)$ is the sigmoid activation function.

The generator $\emph{G}$ and the discriminator $\emph{D}$ are trained simultaneously with the objective function:
\begin{equation}
\begin{aligned}
\Phi _{\text{GAN}} &= \underset{\emph{G}}{\text{min}}\underset{\emph{D}}{\text{max}}\emph{V}(\emph{D},\emph{G}) = \emph{E}_{(\textbf{X}\sim p(\textbf{X}))}\left [ \text{log}(\emph{D}(\textbf{X})) \right ]
\\&+ \emph{E}_{(\textbf{u}^{(\emph{L})}\sim p(\textbf{u}^{(\emph{L})}))}\left [ \text{log}(1-\emph{D}(\emph{G}(\textbf{u}^{(\emph{L})})) \right ]
\label{lossGAN}
\end{aligned}
\end{equation}

The discriminator $\emph{D}$ aims to maximize the probability to give a correct label to both real samples $\textbf{X}$ and fake samples $\emph{G}(\textbf{u}^{(\emph{L})})$, while the generator $\emph{G}$ aims to minimize $\text{log}(1-\emph{D}(\emph{G}(\textbf{u}^{(\emph{L})}))$. Thus, we further define the objective function for the generator $\emph{G}$, which is referred with $\Phi _{\emph{G}}$, and for the discriminator $\emph{D}$, which is referred with $\Phi _{\emph{D}}$.
\begin{equation}
\begin{aligned}
\Phi _{\emph{G}} &= \underset{\emph{G}}{\text{min}}\emph{V}(\emph{D},\emph{G})
= - \emph{E}_{(\textbf{u}^{(\emph{L})}\sim p(\textbf{u}^{(\emph{L})}))}\left [ \text{log}(1-\emph{D}(\emph{G}(\textbf{u}^{(\emph{L})})) \right ]
\label{lossG}
\end{aligned}
\end{equation}

\begin{equation}
\begin{aligned}
\Phi _{\emph{D}} &= \underset{\emph{D}}{\text{min}}\emph{V}(\emph{D},\emph{G})
= -\emph{E}_{(\textbf{X}\sim p(\textbf{X}))}\left [ \text{log}(\emph{D}(\textbf{X})) \right ]
\\&- \emph{E}_{(\textbf{u}^{(\emph{L})}\sim p(\textbf{u}^{(\emph{L})}))}\left [ \text{log}(1-\emph{D}(\emph{G}(\textbf{u}^{(\emph{L})})) \right ]
\label{lossD}
\end{aligned}
\end{equation}

During the training process, the performances of $\emph{G}$ and $\emph{D}$ improve gradually. When $\emph{D}$ cannot distinguish where input samples come from, even though its discrimination ability is very high, it is proved that $\emph{G}$ can generate realistic samples. Moreover, the features are regarded as the most representative at this time. In other words, the optimization process of GAN contributes to examining the quality of features.

\subsubsection{Objective Function of DML-GANR}
\label{sec:Objective Function of DML-GANR:}

With (\ref{lossDML}) and (\ref{lossGAN}), we can formulate the joint objective function of DML-GANR as the weighted sum of the loss of DML and GAN as follows:
\begin{equation}
\begin{aligned}
\Phi &= \Phi _{\text{DML}} + \lambda \Phi _{\text{GAN}} \\&= \underset{\textbf{W}_{\emph{F}}^{(\emph{l})}, \textbf{b}_{\emph{F}}^{(\emph{l})} }{\text{min}} \sum_{l=1}^{L}  [ \emph{S}_{\emph{c}}^{(\emph{l})} - \alpha  \emph{S}_{\emph{b}}^{(\emph{l})} + \gamma  ( \| \textbf{W}_{\emph{F}}^{(\emph{l})}  \|_{\emph{F}}^2 +  \| \textbf{b}_{\emph{F}}^{(\emph{l})}  \|_{\emph{F}}^2  )
 ] \\&+ \lambda \underset{\emph{G}}{\text{min}} \, \underset{\emph{D}}{\text{max}} \, \{ \emph{E}_{(\textbf{X}\sim p(\textbf{X}))} [ \text{log}(\emph{D}(\textbf{X})) ]
\\&+ \emph{E}_{(\textbf{u}^{(\emph{L})}\sim p(\textbf{u}^{(\emph{L})}))} [ \text{log}(1-\emph{D}(\emph{G}(\textbf{u}^{(\emph{L})})))] \}
\label{loss}
\end{aligned}
\end{equation}
\noindent where $\lambda$ is the parameter to balance the contributions of different terms.

\begin{algorithm}[htb]
\caption{DML-GANR.}
\label{alg}
\begin{algorithmic}[1]
\REQUIRE~~\\
Input of HFE: $\textbf{X}$; Input of DML: $\textbf{u}^{(\emph{l})}$ (1 $\leqslant \emph{l} \leqslant \emph{L}$);\\
Input of $\emph{G}$: $\textbf{u}^{(\emph{L})}$; Inputs of $\emph{D}$: $\textbf{X}$, $\emph{G}(\textbf{u}^{(\emph{L})})$;\\
Parameters: $\textbf{W}_{\emph{F}}^{(\emph{l})}$, $\textbf{b}_{\emph{F}}^{(\emph{l})}$, $\textbf{W}_{\emph{G}}^{(\emph{h})}$, $\textbf{b}_{\emph{G}}^{(\emph{h})}$, $\textbf{W}_{\emph{D}}^{(\emph{k})}$, and $\textbf{b}_{\emph{D}}^{(\emph{k})}$(1 $\leqslant \emph{l} \leqslant \emph{L}$, 1 $\leqslant \emph{h} \leqslant \emph{H}$, and 1 $\leqslant \emph{k} \leqslant \emph{K}$).
\ENSURE ~~\\
Output of HFE: $\textbf{u}^{(\emph{l})}$; Outputs of DML: $\textbf{W}_{\emph{F}}^{(\emph{l})}$, $\textbf{b}_{\emph{F}}^{(\emph{l})}$;\\
Output of $\emph{G}$: $\emph{G}(\textbf{u}^{(\emph{L})})$; Outputs of $\emph{D}$: $\emph{D}(\emph{G}(\textbf{u}^{(\emph{L})}))$, $\emph{D}(\textbf{X})$.
\STATE
$\textbf{for}$ number of epoch $\textbf{do}$
\STATE
Sample batch size of images from the training images;
\STATE
Compute the high-level features $\textbf{u}^{(\emph{l})}$ using Eq.(\ref{FC});
\STATE
Calculate the distance metric of features using Eq.(\ref{DMLdistance});
\STATE
Generate images by $\emph{G}$ using Eq.(\ref{HFh});
\STATE
Compute the probability of real and fake data by $\emph{D}$ using Eq.(\ref{HDk});
\STATE
Calculate derivatives of the optimization function of DML using Eq.(\ref{lossDML});
\STATE
Calculate derivatives of the optimization function of $\emph{D}$ using Eq.(\ref{lossD});
\STATE
Calculate derivatives of the optimization function of $\emph{G}$ using Eq.(\ref{lossG});
\STATE
Update parameters of DML $\{\textbf{W}_{\emph{F}}^{(\emph{l})}$, $\textbf{b}_{\emph{F}}^{(\emph{l})}\}$ based on the derivatives via back-propagation;
\STATE
Update parameters of $\emph{D}$ $\{\textbf{W}_{\emph{D}}^{(\emph{k})}$, $\textbf{b}_{\emph{D}}^{(\emph{k})}\}$ based on the derivatives via back-propagation;
\STATE
Update parameters of $\emph{G}$ $\{\textbf{W}_{\emph{G}}^{(\emph{h})}$, $\textbf{b}_{\emph{G}}^{(\emph{h})}\}$ based on the derivatives via back-propagation;
\STATE
$\textbf{end}$ $\textbf{for}$
\end{algorithmic}
\end{algorithm}

\subsection{Optimization of DML-GANR}
\label{sec:Optimization of DML-GANR}

As mentioned above, based on features extracted by multiple FC layers, the multilayer DML is introduced to learn a suitable distance metric among features of the training dataset in a supervised feature learning way, while GAN is utilized to mitigate the overfitting problem and evaluate the quality of the features. In this subsection, we use a back-propagation algorithm to optimize the parameters in multilayer DML and the parameters in the GAN simultaneously.

\subsubsection{Optimization for Multi-Layer DML}
\label{sec:Optimization for Multi-Layer DML}

To solve the optimization problem in (\ref{lossDML}), the gradient descent method is employed to obtain the parameters $\textbf{W}_{\emph{F}}^{(\emph{l})}$ and $\textbf{b}_{\emph{F}}^{(\emph{l})}$ (1$\leqslant \emph{l} \leqslant \emph{L}$). The gradients of the objective function $\Phi _{\text{DML}} $ in (\ref{lossDML}) for the parameters $\textbf{W}_{\emph{F}}^{(\emph{l})}$ and $\textbf{b}_{\emph{F}}^{(\emph{l})}$ are computed as follows:
\begin{equation}
\begin{aligned}
\frac{\partial \Phi _{\text{DML}}(\textbf{W}_{\emph{F}}^{(\emph{l})})}{\partial \textbf{W}_{\emph{F}}^{(\emph{l})}} = \frac{2}{\emph{nt}_1} \sum_{\emph{i=}1}^{\emph{n}} \sum_{\emph{j=}1}^{\emph{t}_1} \emph{P}_{\emph{ij}} \left [ \textbf{M}_{\emph{ij}}^{(\emph{l})} (\textbf{u}^{(\emph{l-}1)}_{\emph{i}})^{\emph{T}} + \textbf{M}_{\emph{ji}}^{(\emph{l})} (\textbf{u}^{(\emph{l-}1)}_{\emph{j}})^{\emph{T}} \right ] \\- \alpha \frac{2}{\emph{nt}_2} \sum_{\emph{i=}1}^{\emph{n}} \sum_{\emph{j=}1}^{\emph{t}_2} \emph{Q}_{\emph{ij}} \left [ \textbf{M}_{\emph{ij}}^{(\emph{l})} (\textbf{u}^{(\emph{l-}1)}_{\emph{i}})^{\emph{T}} + \textbf{M}_{\emph{ji}}^{(\emph{l})} (\textbf{u}^{(\emph{l-}1)}_{\emph{j}})^{\emph{T}} \right ] + 2 \gamma \textbf{W}_{\emph{F}}^{(\emph{l})}
\end{aligned}
\end{equation}

\begin{equation}
\begin{aligned}
\frac{\partial \Phi _{\text{DML}}(\textbf{b}_{\emph{F}}^{(\emph{l})})}{\partial \textbf{b}_{\emph{F}}^{(\emph{l})}} = \frac{2}{\emph{nt}_1} \sum_{\emph{i=}1}^{\emph{n}} \sum_{\emph{j=}1}^{\emph{t}_1} \emph{P}_{\emph{ij}} \left [ \textbf{M}_{\emph{ij}}^{(\emph{l})} + \textbf{M}_{\emph{ji}}^{(\emph{l})} \right ] \\- \alpha \frac{2}{\emph{nt}_2} \sum_{\emph{i=}1}^{\emph{n}} \sum_{\emph{j=}1}^{\emph{t}_2} \emph{Q}_{\emph{ij}} \left [ \textbf{M}_{\emph{ij}}^{(\emph{l})} + \textbf{M}_{\emph{ji}}^{(\emph{l})} \right ] + 2 \gamma \textbf{b}_{\emph{F}}^{(\emph{l})}
\end{aligned}
\end{equation}
\noindent where the updating equations are computed as follows:

\begin{equation}
\begin{aligned}
&\textbf{M}_{\emph{ij}}^{(\emph{l})} = \\& \left \{ \begin{matrix}
\left ( \textbf{u}_{\emph{i}}^{(\emph{L})}- \textbf{u}_{\emph{j}}^{(\emph{L})} \right )\odot {\psi}'(\textbf{z}_{\emph{i}}^{(\emph{L})}), \emph{l}=\emph{L}\\
\left [ \left ( \textbf{u}_{\emph{i}}^{(\emph{l})}- \textbf{u}_{\emph{j}}^{(\emph{l})} \right ) + (\textbf{W}_{\emph{F}}^{(\emph{l+}1)})^{\emph{T}}\textbf{M}_{\emph{ij}}^{(\emph{l+}1)}\right ]\odot {\psi}'(\textbf{z}_{\emph{i}}^{(\emph{l})}), 1\leqslant \emph{l} < \emph{L}
\end{matrix} \right.
\end{aligned}
\end{equation}

\begin{equation}
\begin{aligned}
&\textbf{M}_{\emph{ji}}^{(\emph{l})} = \\& \left \{ \begin{matrix}
\left ( \textbf{u}_{\emph{j}}^{(\emph{L})}- \textbf{u}_{\emph{i}}^{(\emph{L})} \right )\odot {\psi}'(\textbf{z}_{\emph{j}}^{(\emph{L})}), \emph{l}=\emph{L}\\
\left [ \left ( \textbf{u}_{\emph{j}}^{(\emph{l})}- \textbf{u}_{\emph{i}}^{(\emph{l})} \right ) + (\textbf{W}_{\emph{F}}^{(\emph{l+}1)})^{\emph{T}}\textbf{M}_{\emph{ji}}^{(\emph{l+}1)}\right ]\odot {\psi}'(\textbf{z}_{\emph{j}}^{(\emph{l})}), 1\leqslant \emph{l} < \emph{L}
\end{matrix} \right.
\end{aligned}
\end{equation}
\noindent where $\emph{l} = 1,2,...,\emph{L}$, ${\psi}'(\cdot)$ is the derivative of $\psi(\cdot)$, $\odot$ denotes the element-wise multiplication, and $\textbf{z}^{(\emph{l})} = \textbf{W}_{\emph{F}}^{(\emph{l})}\cdot \textbf{u}^{(\emph{l-1})} + \textbf{b}_{\emph{F}}^{(\emph{l})}$.

Therefore, we perform the update on $\textbf{W}_{\emph{F}}^{(\emph{l})}$ and $\textbf{b}_{\emph{F}}^{(\emph{l})}$:
\begin{equation}
\begin{aligned}
\textbf{W}_{\emph{F}}^{(\emph{l})} = \textbf{W}_{\emph{F}}^{(\emph{l})} - \delta \frac{\partial \Phi _{\text{DML}}(\textbf{W}_{\emph{F}}^{(\emph{l})})}{\partial \textbf{W}_{\emph{F}}^{(\emph{l})}}
\end{aligned}
\end{equation}

\begin{equation}
\begin{aligned}
\textbf{b}_{\emph{F}}^{(\emph{l})} = \textbf{b}_{\emph{F}}^{(\emph{l})} - \delta \frac{\partial \Phi _{\text{DML}}(\textbf{b}_{\emph{F}}^{(\emph{l})})}{\partial \textbf{b}_{\emph{F}}^{(\emph{l})}}
\end{aligned}
\end{equation}
\noindent where $\delta $ is the learning rate of DML.

\subsubsection{Optimization for GAN}
\label{sec:Optimization for GAN}

When optimizing the discriminator $\emph{D}$, the parameters of the generator $\emph{G}$ are fixed. Similarly, when optimizing the generator $\emph{G}$, the parameters of the discriminator $\emph{D}$ are fixed. First, we fix $\emph{G}$ and update $\emph{D}$. The partial derivative of weight and bias of $\emph{D}$ at $\emph{k}$th layer are as follows:
\begin{equation}
\begin{aligned}
\frac{\partial \Phi _{\emph{D}}}{\partial \textbf{W}_{\emph{D}}^{(\emph{k})}} = \textbf{T}_{\emph{D}}^{(\emph{k})}( \textbf{H}_{\emph{D}}^{(\emph{k-}1)} )^{\emph{T}}
\end{aligned}
\end{equation}

\begin{equation}
\begin{aligned}
\frac{\partial \Phi _{\emph{D}}}{\partial \textbf{b}_{\emph{D}}^{(\emph{k})}} = \textbf{T}_{\emph{D}}^{(\emph{k})}
\end{aligned}
\end{equation}
\noindent where $\emph{k} = 1,2,...,\emph{K}$.

\begin{equation}
\begin{aligned}
\textbf{T}_{\emph{D}}^{(\emph{k})} = \left\{\begin{matrix}
\left\{\begin{matrix}
-\left (1/(\emph{D}(\textbf{x})) \right )\odot {\sigma}' (\textbf{z}_{\emph{D}}^{(\emph{K})})\\
\left (1/(1-\emph{D}(\emph{G}(\textbf{u}^{(\emph{L})}))) \right )\odot {\sigma}' (\textbf{z}_{\emph{D}}^{(\emph{K})})
\end{matrix}, \emph{k}= \emph{K}\right.\\
\left ( \left (\textbf{W}_{\emph{D}}^{(\emph{k+}1)} \right )^{\emph{T}}\textbf{T}_{\emph{D}}^{(\emph{k+}1)} \right )\odot {\psi}'(\textbf{z}_{\emph{D}}^{(\emph{k})}), 9<\emph{k}<\emph{K}\\
\textbf{T}_{\emph{D}}^{(10)}, \emph{k} = 9\\
\text{up}(\textbf{T}_{\emph{D}}^{(9)}), 7\leqslant \emph{k}\leqslant 8\\
\left ( \emph{R}( \textbf{W}_{\emph{D}}^{(\emph{k+}1)}) {\otimes }'\textbf{T}_{\emph{D}}^{(\emph{k+}1)}\right )\odot {\psi}'(\textbf{z}_{\emph{D}}^{(\emph{k})}), 1\leqslant \emph{k}\leqslant 6
\end{matrix}\right.
\end{aligned}
\end{equation}
\noindent where ${\sigma }'(\cdot )$ is the derivative of $\sigma (\cdot )$. $\text{up}(\cdot)$ is the reversion of the max pooling, and $\emph{R}(\cdot)$ is the rotation of the input matrix by 180 degree. ${\otimes} '$ is the transposed convolution, and $\textbf{z}_{\emph{D}}^{(\emph{k})}$ represents the features of $\emph{k}$th layer that are not activated by the activation function.

We perform the update on $\textbf{W}_{\emph{D}}^{(\emph{k})}$ and $\textbf{b}_{\emph{D}}^{(\emph{k})}$:
\begin{equation}
\begin{aligned}
\textbf{W}_{\emph{D}}^{(\emph{k})} = \textbf{W}_{\emph{D}}^{(\emph{k})} - \beta_1 \frac{\partial \Phi _{\emph{D}}}{\partial \textbf{W}_{\emph{D}}^{(\emph{k})}}
\end{aligned}
\end{equation}

\begin{equation}
\begin{aligned}
\textbf{b}_{\emph{D}}^{(\emph{k})} = \textbf{b}_{\emph{D}}^{(\emph{k})} - \beta_1 \frac{\partial \Phi _{\emph{D}}}{\partial \textbf{b}_{\emph{D}}^{(\emph{k})}}
\end{aligned}
\end{equation}
\noindent where $\beta_1$ is the learning rate of $\emph{D}$.

Then, we fix $\emph{D}$ and update $\emph{G}$. The partial derivative of weight and bias of $\emph{G}$ at $\emph{h}$th layer are as follows:
\begin{equation}
\begin{aligned}
\frac{\partial \Phi _{\emph{G}}}{\partial \textbf{W}_{\emph{G}}^{(\emph{h})}} = \textbf{T}_{\emph{G}}^{(\emph{h})}(\textbf{H}_{\emph{G}}^{(\emph{h-}1)} )^{\emph{T}}
\end{aligned}
\end{equation}

\begin{equation}
\begin{aligned}
\frac{\partial \Phi _{\emph{G}}}{\partial \textbf{b}_{\emph{G}}^{(\emph{h})}} = \textbf{T}_{\emph{G}}^{(\emph{h})}
\end{aligned}
\end{equation}
\noindent where $\emph{h} = 1,2,...,\emph{H}$.

\begin{equation}
\begin{aligned}
\textbf{T}_{\emph{G}}^{(\emph{h})} =
\left\{\begin{matrix}
\left ( \emph{R} (  \textbf{W}_{\emph{D}}^{(1)}  ){\bigotimes }' \textbf{T}_{\emph{D}}^{(1)}\right )\bigodot {\text{Tanh}}'\left (  \textbf{z}_{\emph{G}}^{(\emph{H})}\right ), \emph{h} = \emph{H}\\
\left ( \emph{R}( \textbf{W}_{\emph{G}}^{(\emph{h}+1)} ) {\bigotimes }' \textbf{T}_{\emph{G}}^{(\emph{h}+1)}\right )\bigodot {\varphi }'\left (  \textbf{z}_{\emph{G}}^{(\emph{h})}\right ), 3< \emph{h} < \emph{H}\\
\left (  (\textbf{W}_{\emph{G}}^{(\emph{h}+1)} ) ^{\emph{T}} \textbf{T}_{\emph{G}}^{(\emph{h}+1)}\right )\bigodot {\varphi }'\left (  \textbf{z}_{\emph{G}}^{(\emph{h})}\right ), 1\leqslant \emph{h}\leqslant 3
\end{matrix}\right.
\end{aligned}
\end{equation}
\noindent where ${\varphi }'(\cdot )$ is the derivative of $\varphi (\cdot )$, ${\text{Tanh} }'(\cdot )$ is the derivative of $\text{Tanh} (\cdot )$, and $ \textbf{z}_{\emph{G}}^{(\emph{h})}$ represents features of $\emph{h}$th layer that are not activated by the activation function.

We perform the update on $\textbf{W}_{\emph{G}}^{(\emph{h})}$ and $\textbf{b}_{\emph{G}}^{(\emph{h})}$:
\begin{equation}
\begin{aligned}
\textbf{W}_{\emph{G}}^{(\emph{h})} = \textbf{W}_{\emph{G}}^{(\emph{h})} - \beta _2 \frac{\partial \Phi _{\emph{G}}}{\partial \textbf{W}_{\emph{G}}^{(\emph{h})}}
\end{aligned}
\end{equation}

\begin{equation}
\begin{aligned}
\textbf{b}_{\emph{G}}^{(\emph{h})} = \textbf{b}_{\emph{G}}^{(\emph{h})} - \beta _2 \frac{\partial \Phi _{\emph{G}}}{\partial \textbf{b}_{\emph{G}}^{(\emph{h})}}
\end{aligned}
\end{equation}
\noindent where $ \beta _2$ is the learning rate of $\emph{G}$.

\subsection{Implementation}
\label{sec:Implementation}

The architecture of the DML-GANR is implemented using the Keras deep learning library. We randomly sample a certain number of images from HSR-RSIs for training and use the rest to evaluate the performance of the proposed network. Each training image is randomly cropped to 224 $\times$ 224. To learn the proposed network, adaptive moment estimation (Adam) is used with a batch size of 128 DML samples and 16 GAN samples. For DML, the weight decay is set to 0.0002, and the betas are set to 0.9 and 0.999. The learning rate is set as 0.0002. For GAN, the betas of both generator $\emph{G}$ and discriminator $\emph{D}$ are set to 0.5 and 0.999, respectively. The learning rate of both generator $\emph{G}$ and discriminator $\emph{D}$  is set as 0.0002.

\section{Experiments}
\label{sec:Experiments}

In this section, we evaluate the performance of the proposed DML-GANR method for image retrieval. First, we briefly describe the experimental data sets. Then, we compare the image retrieval performances of our method with related approaches.

\subsection{Experimental Setup}
\label{sec:Experimental Setup}

\begin{figure*}[htb]
\centering\includegraphics[width=7.1 in]{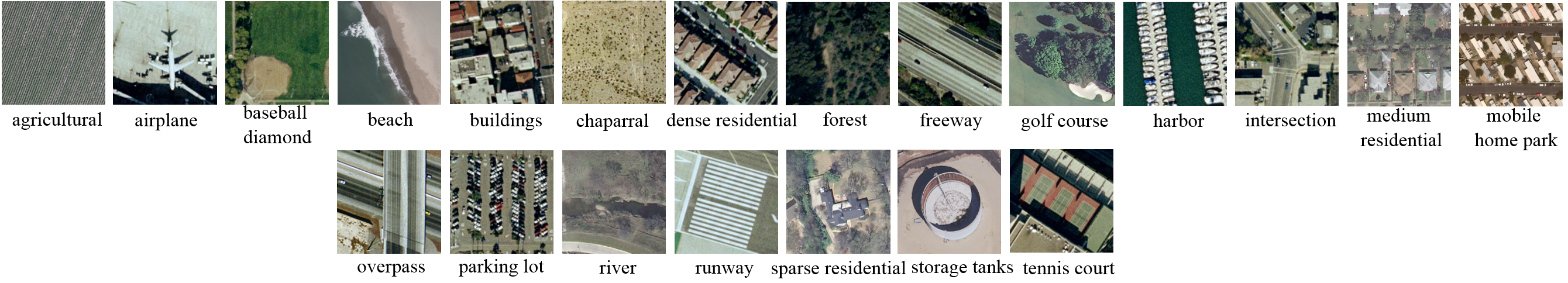}
\caption{One sample image from each class of the UCMD dataset.}\label{Pic_UCMD}
\end{figure*}

\begin{figure*}[htb]
\centering\includegraphics[width=7.1 in]{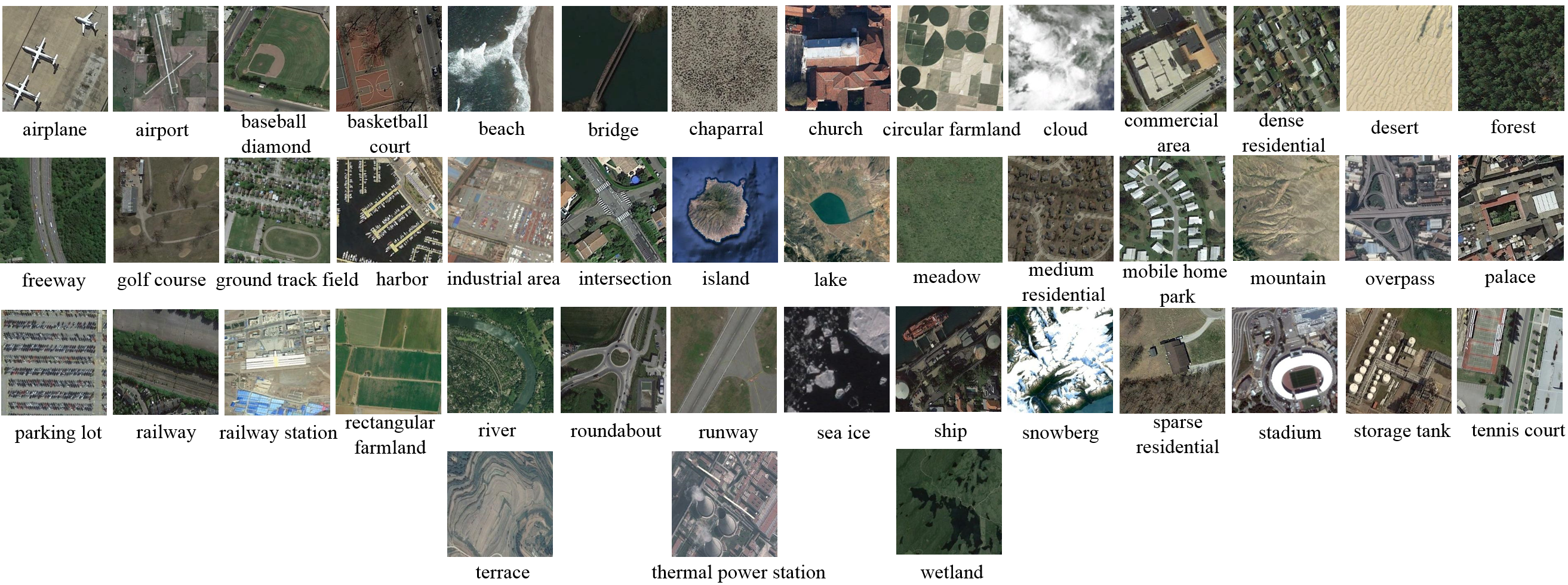}
\caption{One sample image from each class of the NWPU45 dataset.}\label{Pic_NWPU45}
\end{figure*}

\begin{figure*}[htb]
\centering\includegraphics[width=7.1 in]{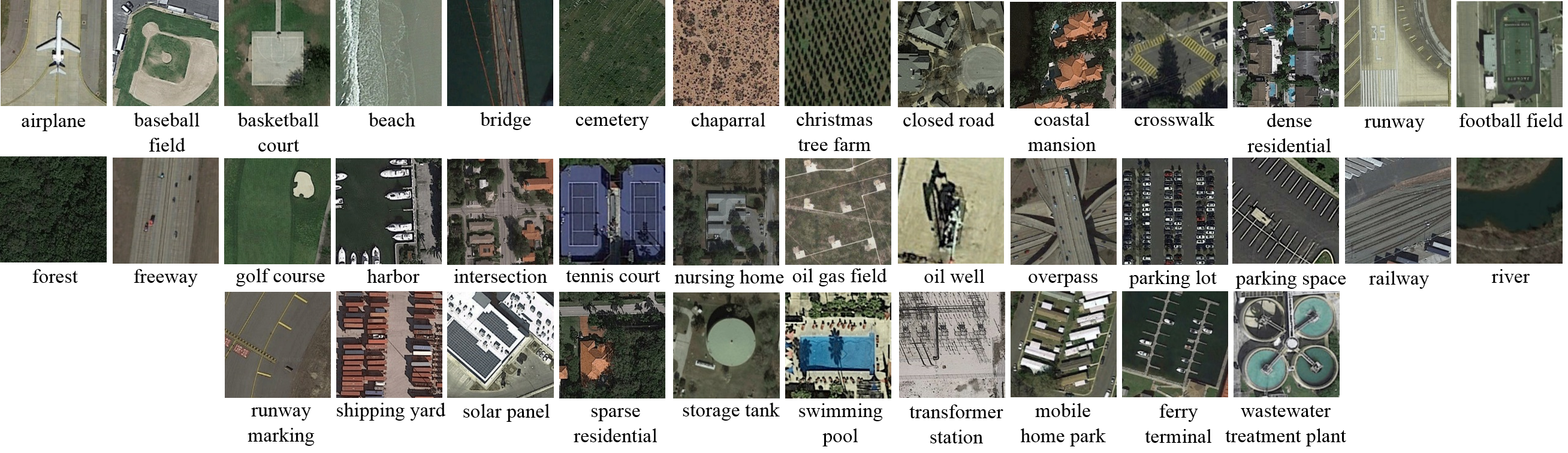}
\caption{One sample image from each class of the PatternNet dataset.}\label{Pic_PatternNet}
\end{figure*}

Three different scene image data sets were used to evaluate the performance of the proposed DML-GANR, which are the UC Merced dataset (UCMD) \cite{yang2010bag}, NWPU-RESISC45 dataset (NWPU45) \cite{Cheng2017Remote}, and PatternNet dataset \cite{zhou2018patternnet:}. We selected one example image from each class and show images from the three datasets in Figs. \ref{Pic_UCMD}-\ref{Pic_PatternNet}.

The UCMD dataset can be downloaded from the U.S. Geological Survey National Map. There are 21 categories with a spatial resolution of 0.3 m. Each class contains 100 images and each image measures 256 $\times$ 256 pixels. As the first publicly available remote sensing evaluation dataset, the UCMD dataset has been used extensively to develop and evaluate HSR-RSI retrieval methods.

The second dataset is the NWPU45 dataset. This dataset has 45 categories and each category contains 700 images. Each image measures 256 $\times$ 256 pixels with a spatial resolution ranging from approximately 0.2 to 30 m. The NWPU45 dataset is a currently large and publicly available dataset, which is constructed by first investigating all scene classes of the existing datasets.

The third dataset is the PatternNet dataset. This dataset has 38 categories and each category contains 800 images. Each image measures 256 $\times$ 256 pixels with a spatial resolution ranging from approximately 0.06 to 4.7 m. The PatternNet dataset is a currently publicly available and high-resolution dataset for HSR-RSI retrieval, which has a large number of images overall making it more suitable for HSR-RSI retrieval approaches based on deep learning.

To evaluate the effectiveness of our method, we conducted extensive experiments on the above three datasets. It should be noted that ResNet-50 requires the input images to have a fixed size of 224$\times $224. As such, all of the images among the three datasets were cropped into 224 $\times $ 224 from an original size 256 $\times $ 256. We first compared our approach in different parts of our proposed DML-GANR method and then further compared our method with other related methods. The mAP, average normalized modified retrieval rank (ANMRR) \cite{manjunath2001color}, precision at $\emph{5}$ and $\emph{50}$ ($\emph{P}@\emph{5}$ and $\emph{P}@\emph{50}$), and precision-recall curves were employed to evaluate the retrieval performance. For mAP, $\emph{P}@\emph{5}$, and $\emph{P}@\emph{50}$, the higher value is better. The mAP can be computed as follows:
\begin{equation}
\begin{aligned}
\text{mAP} = \frac{1}{\emph{Q}}\sum_{\emph{j}=1}^{\emph{Q}}\frac{1}{\emph{m}}\sum_{\emph{i}=1}^{\emph{m}}\text{precision}(\emph{R}_{\emph{j}}^{\emph{i}})
\end{aligned}
\end{equation}

\noindent where $\emph{Q}$ is the number of query images, and $\emph{m}$ is the number of images in the searching image dataset. $\emph{R}_{\emph{j}}^{\emph{i}}$ is the set of ranked results.

For ANMRR, the lower values indicate better performance. The ANMRR can be computed as follows:

\begin{equation}
\begin{aligned}
\text{ANMRR} = \frac{1}{\emph{Q}}\sum_{\emph{q}=1}^{\emph{Q}} \frac{\text{AR}(\emph{q})-0.5[1+\text{NG}(\emph{q})]}{1.25\emph{K}(\emph{q})-0.5[1+\text{NG}(\emph{q})]}
\end{aligned}
\end{equation}

\noindent where $\text{AR}(\emph{q}) = \frac{1}{\text{NG}(\emph{q})}\sum_{\emph{k}=1}^{\text{NG}(\emph{q})}\emph{R}(\emph{k})$ is the average rank. $\text{NG}(\emph{q})$ is the number of the ground truth set for a query image $\emph{q}$. $\emph{K}(\emph{q}) = \text{min}\{4\text{NG}(\emph{q}), 2 \text{GTM}\}$ is the relevant rank for $\emph{q}$, where GTM is the maximum of $\text{NG}(\emph{q})$ for all queries. $\emph{R}(\emph{k})$ is the retrieved rank of $\emph{k}$th image. The value of $\emph{R}(\emph{k})$ is $1.25\emph{K}(\emph{q})$ when $\emph{R}(\emph{k})$ is larger than $\emph{K}(\emph{q})$.

\subsection{Validation of Retrieval Performance in Each Part of the DML-GANR}
\label{sec:Validation of Retrieval Performance in Each Part of DML-GANR}

As illustrated in the above sections, our DML-GANR method has three parts: HFE, DML, and GAN. The performance of the three parts of DML-GANR was validated in this section. In HFE, we extracted 2,048-dimensional average pooling features from the pretrained ResNet-50 and applied three FC layers to reduce the dimension of the feature to 1,024. Multilayer DML was employed at each FC layer. The feature extracted from the last FC layer was received as the input of the generator \emph{G} in GAN. We randomly chose 70\% of the labeled images from each class of dataset as the training dataset. The remaining 30\% of images were selected as the query images for the retrieval performance evaluation, which were also called the test dataset. Each query image searched the images from the test dataset instead of the training dataset and the entire dataset. This is because the features of the images in the training dataset were trained and are representative. Searching the images both from the training dataset and the entire dataset will make the retrieval performance inaccurate. In the following experimental results, we took each test image as a query image, where the mAP, ANMRR, $\emph{P}@\emph{5}$, and $\emph{P}@\emph{50}$ are the average of all the queries.

\subsubsection{Retrieval Performance in the HFE}
\label{sec:Retrieval Performance in the HFE}

\renewcommand\arraystretch{1.25}
\begin{table}[htbp]
\caption{The Architecture of HFE.}
\centering{}%
\begin{tabular}{c|c|c c| c c c c}
\hline
Nets & No. & \multicolumn{2}{c|}{Layer} & BN & Str & Pad & Act \\
\hline
CNN & conv1 & \multicolumn{2}{c|}{7$\times$7$\times$64} &Y & 2& 3& ReLU\\
\cline{2-8}
&$\text{conv2}\_\text{x}$ & \multicolumn{2}{c|}{Maxpool.3$\times$3}& N& 2& 1& -\\
& &1$\times$1 $\times$ 64 & \multirow{3}{*}{$\times$3} & Y & 1& 1& ReLU\\
& &3$\times$3 $\times$ 64 & &Y & 1/2 &1& ReLU\\
& &1$\times$1 $\times$ 256 & &Y &1 &1& ReLU\\
\cline{2-8}
& $\text{conv3}\_\text{x}$ &1$\times$1 $\times$ 128 & \multirow{3}{*}{$\times$4} & Y & 1& 1& ReLU\\
& &3$\times$3 $\times$ 128 & &Y & 1/2 &1& ReLU\\
& &1$\times$1 $\times$ 512 & &Y &1 &1& ReLU\\
\cline{2-8}
& $\text{conv4}\_\text{x}$ &1$\times$1 $\times$ 256 & \multirow{3}{*}{$\times$6} & Y & 1& 1& ReLU\\
& &3$\times$3 $\times$ 256 & &Y & 1/2 &1& ReLU\\
& &1$\times$1 $\times$ 256 & &Y &1 &1& ReLU\\
\cline{2-8}
& $\text{conv5}\_\text{x}$ &1$\times$1 $\times$ 512 & \multirow{3}{*}{$\times$3} & Y & 1& 1& ReLU\\
& &3$\times$3 $\times$ 512 & &Y & 1/2 &1& ReLU\\
& &1$\times$1 $\times$ 2,048 & &Y &1 &1& ReLU\\
\cline{2-8}
& Avepool & \multicolumn{2}{c|}{7$\times$7} & N &1 &0& -\\
\hline
FC & FC	&\multicolumn{2}{c|}{2,048$\times$1,024}&N&1&0	&LReLU\\
& FC	&\multicolumn{2}{c|}{1,024$\times$1,024}&N&1&0	&LReLU\\
& FC	&\multicolumn{2}{c|}{1,024$\times$1,024}&N&1&0	&LReLU\\
\hline
\end{tabular}
\label{AHFE}
\begin{tablenotes}
    \item $\emph{Y is Yes, and N is No. Str is Stride, and Act is activation function}$.
\end{tablenotes}
\end{table}

\renewcommand\arraystretch{1.25}
\begin{table*}[htbp]
\caption{The Results of Convolutional Layers of HFE on UCMD, NWPU45, and PatternNet.}
\centering{}
\begin{tabular}{c|c c c c| c c c c| c c c c}
\hline
Features & \multicolumn{4}{c|}{UCMD} & \multicolumn{4}{c|}{NWPU45} & \multicolumn{4}{c}{PatternNet}\\
\cline{2-13}
&ANMRR & mAP & $\emph{P}@\emph{5}$ &$\emph{P}@\emph{50}$ &ANMRR & mAP & $\emph{P}@\emph{5}$ & $\emph{P}@\emph{50}$& ANMRR & mAP & $\emph{P}@\emph{5}$ &$\emph{P}@\emph{50}$\\
\hline
NasnetMobile	&0.7552&	0.1196&	0.3632&	0.1236 &0.8718	&0.0475	&0.2963	&0.0995&0.7731	&0.1068&	0.4736	&0.2318\\
MobileNet&0.5524&	0.2911	&0.6603	&0.2321&0.7636&	0.1039&	0.4903	&0.2346&0.3530	&0.4875	&0.8364	&0.6089	\\
VGG-16	&0.4300	&0.4098	&0.7514	&0.3053& 0.6341&	0.2021&	0.6719	&0.4015&0.3897	&0.4775	&0.9216	&0.7555\\
VGG-19	& 0.4270	&0.4107	&0.7524	&0.3090 &0.6427&	0.1929&	0.6627	&0.3902 & 0.3931	&0.4683	&0.9117	&0.7421 \\
ResNet-50	&\textbf{0.3826}	&\textbf{0.4435}	&\textbf{0.7937}	&\textbf{0.3270} & \textbf{0.5403}&	\textbf{0.2762}&	\textbf{0.7456}	&\textbf{0.4923} & \textbf{0.2571}	&\textbf{0.6083}	&\textbf{0.9565}&	\textbf{0.8465}\\
DenseNet-121	&0.7371&	0.1439&	0.4476&	0.1386	&	0.8349&	0.0701&	0.4141	&0.1683 &	0.6852	&0.1832	&0.7070	&0.3920\\
DenseNet-169	& 0.5816	&0.2647	&0.6200	&0.2192 &	0.7548&	0.1130&	0.5453	&0.2645 &	0.5233	&0.3341	&0.8506	&0.6003\\
DenseNet-201	& 0.7243&	0.1528	&0.4549	&0.1429 &0.8429&	0.0671&	0.4057	&0.1631&	0.7198	&0.1602	&0.6911	&0.3706\\
\hline
\end{tabular}
\begin{tablenotes}
    \item $\emph{For ANMRR, lower values indicate better performance, while for mAP and P@k, larger is better}$.
\end{tablenotes}
\label{P_CNN}
\end{table*}

\renewcommand\arraystretch{1.25}
\begin{table*}[htbp]
\caption{The Results of FC Layers of HFE on UCMD, NWPU45, and PatternNet.}
\centering{}
\begin{tabular}{c|c c c c| c c c c| c c c c}
\hline
Features & \multicolumn{4}{c|}{UCMD} & \multicolumn{4}{c|}{NWPU45} & \multicolumn{4}{c}{PatternNet}\\
\cline{2-13}
&ANMRR & mAP & $\emph{P}@\emph{5}$ &$\emph{P}@\emph{50}$ &ANMRR & mAP & $\emph{P}@\emph{5}$ & $\emph{P}@\emph{50}$& ANMRR & mAP & $\emph{P}@\emph{5}$ &$\emph{P}@\emph{50}$\\
\hline
2,048(ResNet-50)&	0.3826&	0.4435&	0.7937&	0.3270 &	0.5403&	0.2762	&0.7456	&0.4923 &	0.2571&	0.6083&	0.9565&	0.8465\\
1,024	&\textbf{0.3062}&	\textbf{0.5356}&	\textbf{0.8410}&	\textbf{0.3784} &	\textbf{0.4479}&	\textbf{0.3401}	&\textbf{0.7946}	&\textbf{0.5672} &	\textbf{0.1729}	&\textbf{0.7259}	&\textbf{0.9715}&	\textbf{0.9032}\\
512&	0.3237&	0.5157	&0.8311	&0.3647 &	0.4954&	0.3241&	0.7817	&0.5500 &0.1912&	0.7033&	0.9671&	0.8893\\
256&	0.3223&	0.5150	&0.8305&	0.3660 & 0.5010&	0.3172&	0.7668	&0.5373 &	0.1737&	0.7199&	0.9672&	0.8933\\
1024+1024&	\textbf{0.2695}&	\textbf{0.5712}&	\textbf{0.8419}&	\textbf{0.4012} &	\textbf{0.4513}&	\textbf{0.3637}&	\textbf{0.7963}&	\textbf{0.5717} &	\textbf{0.1557}&	\textbf{0.7423}&\textbf{	0.9726}&	\textbf{0.9086}\\
1024+512&	0.2837&	0.5529	&0.8232&	0.3899  &	0.4557&	0.3546&	0.7929&	0.5745 &	0.1555&	0.7429&	0.9704&	0.9055\\
1024+256&	0.2806&	0.5575&	0.8343	&0.3919 &	0.4628&	0.3471&	0.7719&	0.5552 &0.1589&	0.7341&	0.9678&	0.8991\\
1024+1024+1024&	\textbf{0.2358}&	\textbf{0.6045}&	\textbf{0.8562}	&\textbf{0.4189} &	\textbf{0.3850}&	\textbf{0.4233}&	\textbf{0.8202}&	\textbf{0.6314} &	\textbf{0.1421}&	\textbf{0.7594}&	\textbf{0.9730}&	\textbf{0.9124}\\
1024+512+512&	0.2600&	0.5787	&0.8460&	0.4034 &	0.4016&	0.4044&	0.8039&	0.6110 &	0.1593	&0.7393&	0.9675&	0.9013 \\
1024+512+256&	0.2623&	0.5668&	0.8235&	0.4023 &	0.4049&	0.4010&	0.7960&	0.6015 &	0.1488&	0.7445&	0.9660&	0.9012\\
\hline
\end{tabular}
\begin{tablenotes}
    \item $\emph{For ANMRR, lower values indicate better performance, while for mAP and P@k, larger is better}$.
\end{tablenotes}
\label{P_FC}
\end{table*}

The architecture of HFE is shown in Table \ref{AHFE}. In the convolutional layers of HFE (i.e., pretrained ResNet-50), we sequentially fed our training images into ResNet-50 and extracted the 2,048-dimensional feature from the average pooling layer. The weights and bias of ResNet-50 were fixed and did not need to be updated. To validate the retrieval performance in the first layer, we used eight methods to retrieve images, namely NasnetMobile, MobileNet, VGG-16, VGG-19, ResNet-50, DenseNet-121, DenseNet-169, and DenseNet-201. Similarly, the weights and bias of NasnetMobile, MobileNet, VGG-16, VGG-19, ResNet-50, DenseNet-121, DenseNet-169, and DenseNet-201 were fixed.

The mAP, ANMRR, $\emph{P}@\emph{5}$, and $\emph{P}@\emph{50}$ of the query images of the three datasets are listed in Table \ref{P_CNN}. From Table \ref{P_CNN}, the best performance of various pretrained CNNs as achieved by ResNet-50, demonstrating that deeper networks tend to achieve higher retrieval performance than shallower networks (i.e., VGG-16, VGG-19, MobileNet, and NasnetMobile). However, when the depth of networks was larger than 100 (i.e., DenseNet-121, DenseNet-169, DenseNet-201), the retrieval performances were lower than ResNet-50. Therefore, we chose ResNet-50 as the pretrained CNN since the features extracted by ResNet-50 were more representative and achieved good performance.

In the FC layers of HFE, to find the optimal combinations of FC layers, we evaluated the network by varying the number of neurons in each FC layer: 256, 512, and 1,024 for all three datasets. The 2,048-dimensional features extracted from the pretrained ResNet-50 were fed into FC layers. The FC layers were trained with a softmax classifier, while the weights and bias of FC layers were updated using a back-propagation algorithm. The mAP, ANMRR, $\emph{P}@\emph{5}$, and $\emph{P}@\emph{50}$ were used to evaluate different retrieval results. The retrieval results of the three datasets are shown in Table. \ref{P_FC}. For the three datasets, the three-FC layer achieved the best performance among all variations. The results show that 1,024 was the optimal number of neurons when the one-FC layer was wired to the pretrained CNN. The optimal combination for the two-FC layer was 1024+1024, while 1024+1024+1024 was the optimal combination for the three-FC layer. The best performances of one-FC layer, two-FC layer, and three-FC layer were selected to feed into the DML to compare their performance in Section \ref{sec:Retrieval Performance in the Multi-Layer DML}.

\subsubsection{Retrieval Performance in the Multilayer DML}
\label{sec:Retrieval Performance in the Multi-Layer DML}

\begin{figure}[htb]
\centering
{
\subfigure[] {\includegraphics[width=2.3 in,clip]{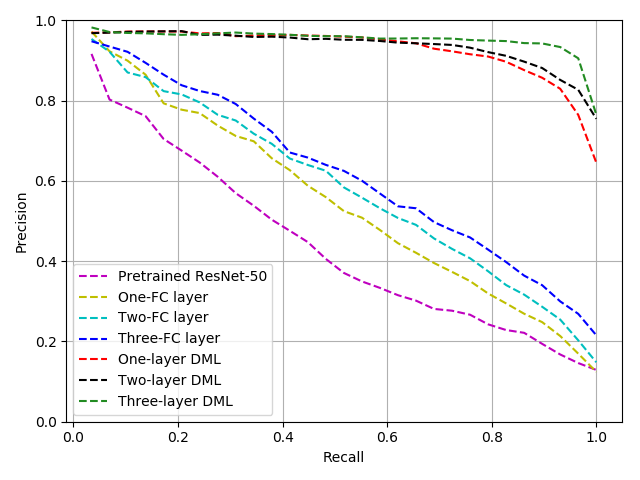}}
\subfigure[] {\includegraphics[width=2.3 in,clip]{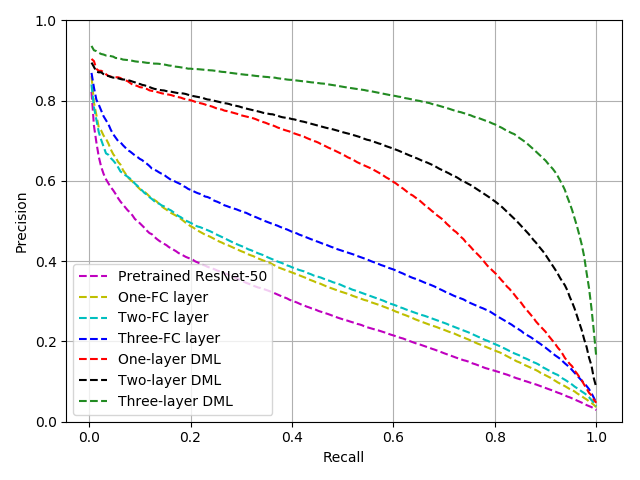}}
\subfigure[] {\includegraphics[width=2.3 in,clip]{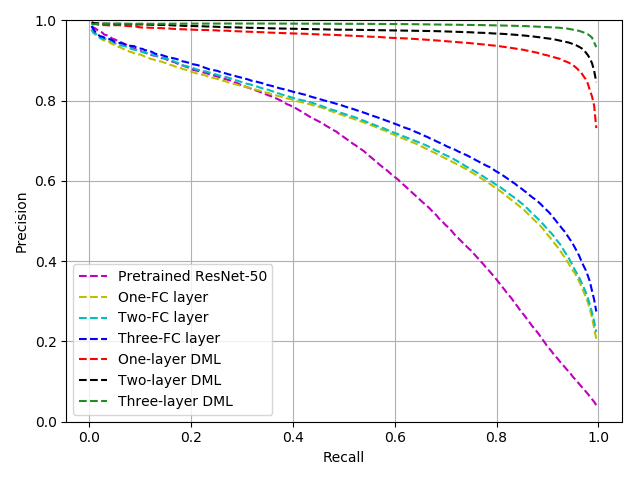}}
}
\caption{The precision-recall curves of pretrained ResNet-50, FC layers, and multi-layer DML on the three datasets. (a) UCMD (b) NWPU45 (c) PatternNet.}\label{PR_MDML}
\end{figure}

\renewcommand\arraystretch{1.25}
\begin{table*}[htbp]
\caption{The Results of Pretrained ResNet-50, FC Layers, and Multilayer DML on UCMD, NWPU45, and PatternNet.}
\centering{}%
\begin{tabular}{c|c|c c c c| c c c c| c c c c}
\hline
Layer&Layer & \multicolumn{4}{c|}{UCMD} & \multicolumn{4}{c|}{NWPU45} & \multicolumn{4}{c}{PatternNet}\\
\cline{3-14}
Name&Number&ANMRR & mAP & $\emph{P}@\emph{5}$ &$\emph{P}@\emph{50}$ &ANMRR & mAP & $\emph{P}@\emph{5}$ & $\emph{P}@\emph{50}$& ANMRR & mAP & $\emph{P}@\emph{5}$ &$\emph{P}@\emph{50}$\\
\hline
ResNet-50 &2048	&0.3826&0.4435&	0.7937&	0.3270 &0.5403&	0.2762&	0.7456&	0.4923 &0.2571	&0.6083&	0.9565&	0.8465\\
\hline
FC layers&One&	0.3062&	0.5356&	0.8410&	0.3784 &0.4479&	0.3401&	0.7946&	0.5672 &0.1729&	0.7259&	0.9715&	0.9032\\
&Two&	0.2695&	0.5712&	0.8419&	0.4012 &	0.4513&	0.3637&	0.7963&	0.5717&	0.1557&	0.7423	&0.9726&	0.9086\\
&Three&	0.2358&	0.6045&	0.8562&	0.4189 &	0.3850&	0.4233&	0.8202	&0.6314 &	0.1421&	0.7594&	0.9730&	0.9124\\
\hline
Multilayer &One&	0.0483&	0.8962&	0.9641&	0.5591 &0.2255&	0.6536	&0.8960&	0.8132 &	0.0220	&0.9469	&0.9914&	0.9788\\
DML &Two&	0.0391&	0.9140&	0.9673&	0.5661 &0.1647&	0.7411&	0.9207	&0.8646 &0.0120	&0.9718	&0.9954	&0.9898\\
&Three&	\textbf{0.0284}&	\textbf{0.9343}&	\textbf{0.9727}&	\textbf{0.5765} &\textbf{0.1132}&	\textbf{0.8095}&	\textbf{0.9345}&\textbf{0.8937} &	\textbf{0.0042}&	\textbf{0.9885}&	\textbf{0.9962}&	\textbf{0.9936}\\
\hline
\end{tabular}
\begin{tablenotes}
    \item $\emph{For ANMRR, lower values indicate better performance, while for mAP and P@k, larger is better}$.
\end{tablenotes}
\label{P_MDML}
\end{table*}

\begin{figure}[htb]
\centering\includegraphics[width=3.2 in]{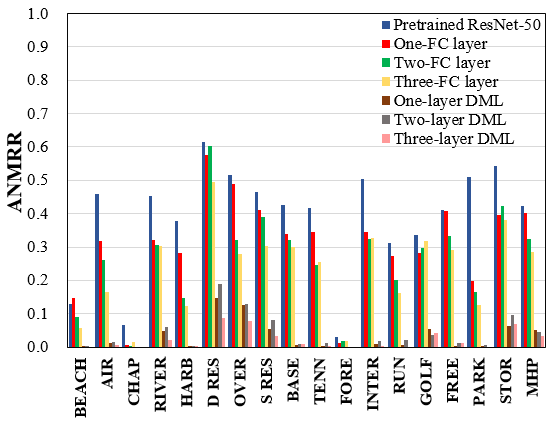}
\caption{The ANMRR for each class of the UCMD dataset extracted by pretrained ResNet-50, FC layers and multilayer DML, respectively.}\label{ANMRR_DMLU}
\end{figure}

\begin{figure}[htb]
\centering\includegraphics[width=3.2 in]{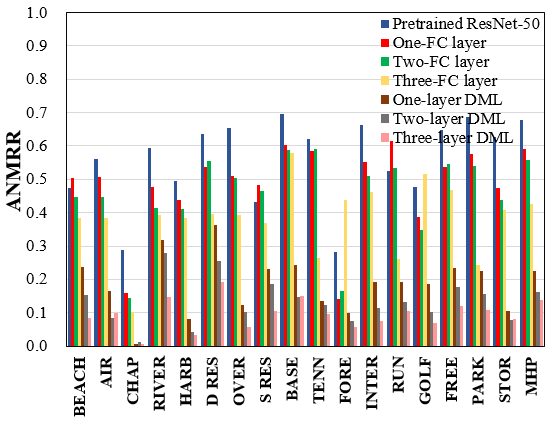}
\caption{The ANMRR for each class of the NWPU45 dataset extracted by pretrained ResNet-50, FC layers and multilayer DML, respectively.}\label{ANMRR_DMLN}
\end{figure}

\begin{figure}[htb]
\centering\includegraphics[width=3.2 in]{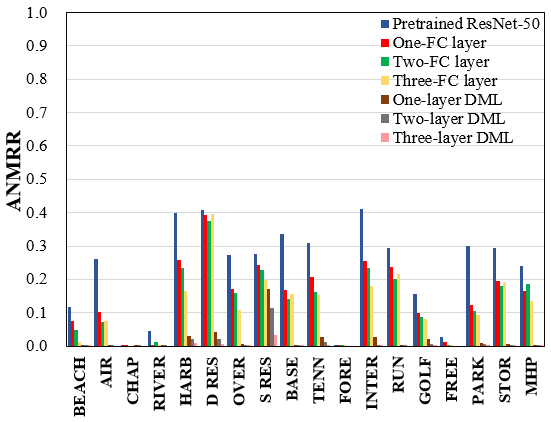}
\caption{The ANMRR for each class of the PatternNet dataset extracted by pretrained ResNet-50, FC layers and multilayer DML, respectively.}\label{ANMRR_DMLP}
\end{figure}

In the multilayer DML part, we fed neurons of each FC layer into the DML to maximize the interclass distances and minimize the intraclass distances. We compared the results of DML with pretrained ResNet-50 and FC layers. The results of the three-layer DML (1024+1024+1024) were compared with those of the one-layer DML (1024) and two-layer DML (1024+1024). The mAP, ANMRR, $\emph{P}@\emph{5}$, and $\emph{P}@\emph{50}$ were used to evaluate different retrieval results. Fig. \ref{PR_MDML} shows the precision-recall curves for features of pretrained ResNet-50, FC layers, and multilayer DML. We selected 18 categories that overlap in the three datasets and calculated the ANMRR for each class extracted by pretrained ResNet-50, FC layers and multilayer DML. The ANMRR for each class is shown in Figs. \ref{ANMRR_DMLU}-\ref{ANMRR_DMLP}. In Table \ref{P_MDML}, the ANMRR, mAP, $\emph{P}@\emph{5}$, and $\emph{P}@\emph{50}$ are shown.

For the UCMD dataset, the mAPs were 44.35\%, 53.56\%, 57.12\%, 60.45\%, 89.62\%, 91.40\% and 93.43\%, corresponding to the retrieval results in pretrained ResNet-50, one-FC layer, two-FC layer, three-FC layer, one-layer DML, two-layer DML and three-layer DML, respectively. The three-layer DML significantly outperformed the pretrained ResNet-50 and three-FC layer by 49.08\% and 32.98\%, respectively. The three-layer DML outperformed the one-layer DML and two-layer DML by 3.81\% and 2.03\%, respectively. For the NWPU45 dataset, the mAPs were 27.62\%, 34.01\%, 36.37\%, 42.33\%, 65.36\%, 74.11\% and 80.95\%, respectively. The three-layer DML significantly outperformed the pretrained ResNet-50 and three-FC layer by 53.33\% and 38.62\%, respectively. The three-layer DML outperformed the one-layer DML and two-layer DML by 15.59\% and 6.84\%, respectively. For the PatternNet dataset, the mAPs were 60.83\%, 72.59\%, 74.23\%, 75.94\%, 94.69\%, 97.18\% and 98.56\%, respectively. The three-layer DML significantly outperformed the pretrained ResNet-50 and three-FC layer by 37.73\% and 22.62\%, respectively. The three-layer DML outperformed the one-layer DML and two-layer DML by 4.16\% and 1.38\%, respectively. The $\emph{P}@\emph{5}$ and $\emph{P}@\emph{50}$ of the three-layer DML were higher than those of the pretrained ResNet-50, FC layers, one-layer DML, and two-layer DML for the three datasets. For ANMRR, lower values indicate better performance \cite{zhou2018patternnet:}. The ANMRR of the three-layer DML was lower than that of the pretrained ResNet-50, FC layers, one-layer DML, and two-layer DML for the three datasets. As shown in Figs. \ref{ANMRR_DMLU}-\ref{ANMRR_DMLP}, for the ANMRR of each class, the retrieval results of the multilayer DML had high precision for most categories. The performances of the multilayer DML significantly outperformed those of the pretrained ResNet-50 and FC layers. For the three datasets, some categories were relatively easy to recognize, such as chaparral, forest, and beach. For the NWPU45 dataset, the ANMRRs of the beach were 0.4744, 0.5036, 0.4461, 0.3829, 0.2383, 0.1545, and 0.0844, corresponding to the pretrained ResNet-50, one-FC layer, two-FC layer, three-FC layer, one-layer DML, two-layer DML and three-layer DML, respectively. However, sparse residential and dense residential were relatively difficult to distinguish for pretrained ResNet-50 and FC layers. For the UCMD dataset, the ANMRRs of dense residential areas were 0.6139, 0.5762, 0.6027, 0.4947, 0.1468, 0.1888, and 0.0882. For the PatternNet dataset, the ANMRRs of the multilayer DML were relatively lower than those of the pretrained ResNet-50 and FC layers, even equaling 0, such as airplane, and chaparral.

\subsubsection{Generative Performance Comparisons of features in the GAN}
\label{sec:Generative Performance Comparisons of features in the GAN}

\begin{figure}[htb]
\centering\includegraphics[width=3.5 in]{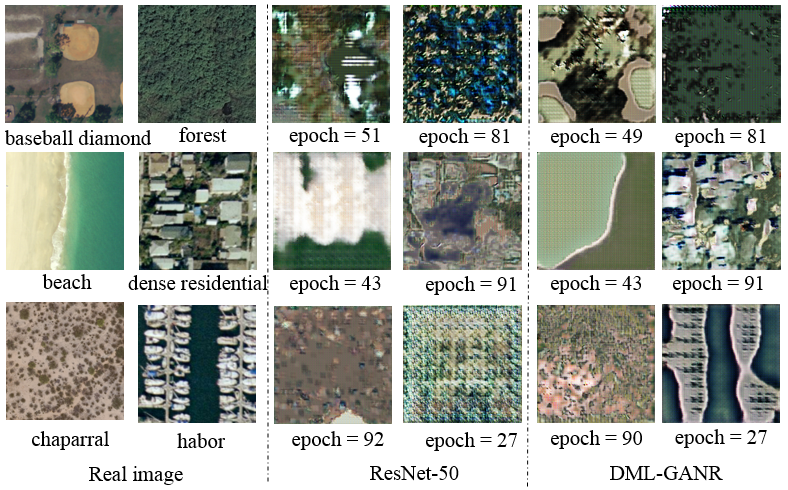}
\caption{Images generated by GAN with the input extracted from ResNet-50 and DML-GANR based on the UCMD dataset.}\label{GANU}
\end{figure}

\begin{figure}[htb]
\centering\includegraphics[width=3.5 in]{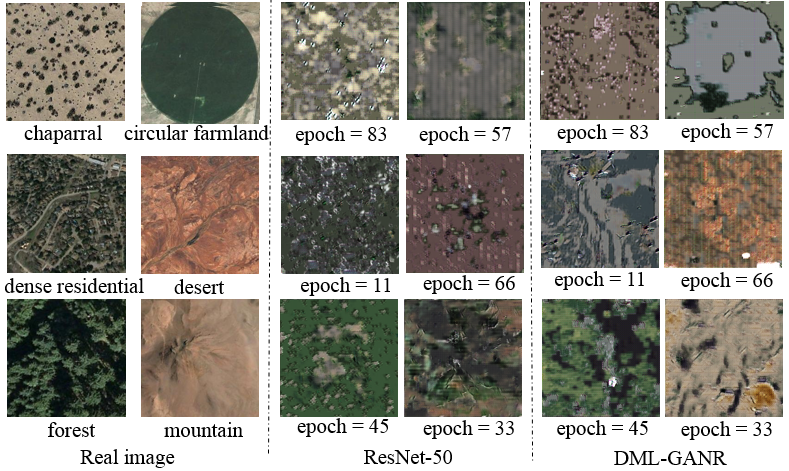}
\caption{Images generated by GAN with the input extracted from ResNet-50 and DML-GANR based on the NWPU45 dataset.}\label{GANN}
\end{figure}

\begin{figure}[htb]
\centering\includegraphics[width=3.5 in]{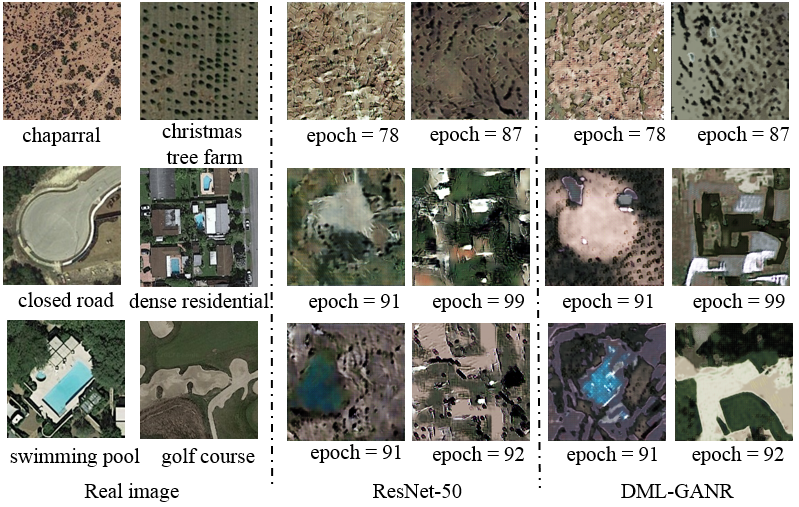}
\caption{Images generated by GAN with the input extracted from ResNet-50 and DML-GANR based on the PatternNet dataset.}\label{GANP}
\end{figure}

In Section \ref{sec:High-Level Feature Extraction (HFE)}, we extracted high-level features from HFE. The features of the last layer of the HFE were saved as the input of the generator $\emph{G}$ in the GAN. The size of real images of the three datasets was 256$\times$ 256 uniformly, and the size of fake images generated by $\emph{G}$ was 256 $\times$ 256 as well. The input images were normalized into the range [-1,1]. The discriminator $\emph{D}$ received real images and fake images as the input and determined the probability of the image coming from the true data rather than simulation by $\emph{G}$. The networks of $\emph{G}$ and $\emph{D}$ are shown in Table \ref{AGAN}. Batch normalization (BN) was used in both $\emph{G}$ and $\emph{D}$ between the convolutional layer and the activation function to improve the performance and stability of GAN \cite{ioffe2015batch}. The activation function of layers 1-16 in $\emph{G}$ was ReLU and that of layer 17 as Tanh. The activation function of layers 1-6 and 10 in $\emph{D}$ was LReLU and layer 11 as sigmoid.
\renewcommand\arraystretch{1.25}
\begin{table}[htbp]
\caption{The Architecture of GAN.}
\centering{}
\begin{tabular}{c|c c c c c c}
\hline
 & No. & Layer & BN & Str & Pad & Act \\
\hline
\multirow{17}{*}{$\emph{G}$}& $\emph{G}1$ & FC.512$\times$1,024 &N & 1& 0& ReLU\\
& $\emph{G}$2 & FC.1,024$\times$4,096 &N & 1& 0& ReLU\\
& $\emph{G}$3 & FC.4,096$\times$(512$\times$4$\times$4) &N & 1& 0& ReLU\\
& $\emph{G}$4 & Conv.4$\times$4$\times$256 &Y & 2& 1& ReLU\\
& $\emph{G}$5 & Conv.3$\times$3$\times$256 &Y & 1& 1& ReLU\\
& $\emph{G}$6 & Conv.3$\times$3$\times$256 &Y & 1& 1& ReLU\\
& $\emph{G}$7 & Conv.4$\times$4$\times$128 &Y & 2& 1& ReLU\\
& $\emph{G}$8 & Conv.3$\times$3$\times$128 &Y & 1& 1& ReLU\\
& $\emph{G}$9 & Conv.3$\times$3$\times$128 &Y & 1& 1& ReLU\\
& $\emph{G}$10 & Conv.4$\times$4$\times$64 &Y & 2& 1& ReLU\\
& $\emph{G}$11 & Conv.3$\times$3$\times$64 &Y & 1& 1& ReLU\\
& $\emph{G}$12 & Conv.3$\times$3$\times$64 &Y & 1& 1& ReLU\\
& $\emph{G}$13 & Conv.4$\times$4$\times$32 &Y & 2& 1& ReLU\\
& $\emph{G}$14 & Conv.3$\times$3$\times$32 &Y & 1& 1& ReLU\\
& $\emph{G}$15 & Conv.3$\times$3$\times$32 &Y & 1& 1& ReLU\\
& $\emph{G}$16 & Conv.4$\times$4$\times$16 &Y & 2& 1& ReLU\\
& $\emph{G}$17 & Conv.4$\times$4$\times$3 &N & 2& 1& Tanh\\
\hline
\multirow{11}{*}{$\emph{D}$} & $\emph{D}1$  & Conv.4$\times$4$\times$16 &N & 2& 1& LReLU\\
& $\emph{D}2$ & Conv.4$\times$4$\times$32 &Y & 2& 1& LReLU\\
& $\emph{D}3$ & Conv.4$\times$4$\times$64 &Y & 2& 1& LReLU\\
& $\emph{D}4$ & Conv.4$\times$4$\times$128 &Y & 2& 1& LReLU\\
& $\emph{D}5$ & Conv.4$\times$4$\times$256 &Y & 2& 1& LReLU\\
& $\emph{D}6$ & Conv.4$\times$4$\times$512 &Y & 2& 1& LReLU\\
& $\emph{D}7$ & Maxpool.4$\times$4(To $\emph{D}4$) &N & 4& 0& -\\
& $\emph{D}8$ & Maxpool.2$\times$2(To $\emph{D}5$) &N & 2& 0& -\\
& $\emph{D}9$ & Cat.($\emph{D}6$,$\emph{D}7$,$\emph{D}8$) &N & 1& 0& -\\
& $\emph{D}10$ & FC.(896$\times$4$\times$4)$\times$1024 &N & 1& 0& LReLU\\
& $\emph{D}11$ & FC.1,024$\times$1 &N & 1& 0& Sigmoid\\
\hline
\end{tabular}
\label{AGAN}
\begin{tablenotes}
    \item $\emph{Y is Yes, and N is No. Str is Stride, and Act is activation function}$.
\end{tablenotes}
\end{table}
We enumerate some of the generated images produced by $\emph{G}$ trained on the three datasets in Fig. \ref{GANU}-\ref{GANP}. In Fig. \ref{GANU}-\ref{GANP}, it is demonstrated that features constrained by the DML of our DML-GANR method encourage the fake image to be more similar to the real image than features extracted from the pretrained CNN. For the same epoch, fake images generated from DML-GANR were more realistic to the real image than those generated from ResNet-50. For the UCMD dataset, DML-GANR generated more realistic images of the baseball diamond and chaparral than ResNet-50 in an earlier epoch. It was also demonstrated that the features of DML-GANR were representative and more powerful than those extracted from ResNet-50.


\subsubsection{Effectiveness of the Parameter Settings}
\label{sec:Effectiveness of the Parameter Settings}
\begin{figure}[htb]
\centering
{
\subfigure[] {\includegraphics[width=1.7 in,clip]{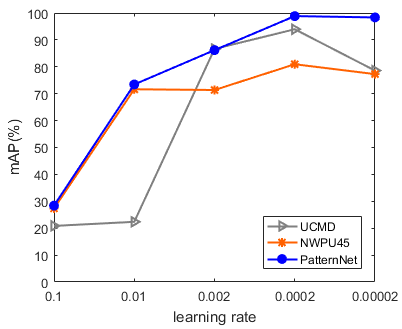}}
\subfigure[] {\includegraphics[width=1.7 in,clip]{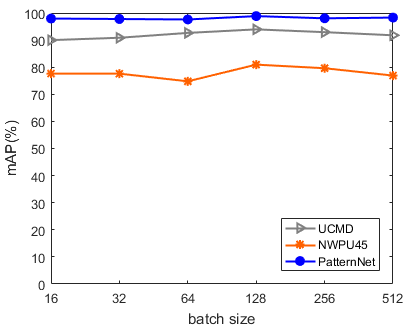}}
}
\caption{The results of different parameter settings with the learning rate and the batch size on the three datasets. (a) the learning rate (b) the batch size.}\label{Parameter Settings}
\end{figure}
To find the optimal parameters of the proposed network, we evaluated the network by varying the number of the learning rate and the batch size. The learning rate is selected from the set \{0.1, 0.01, 0.002, 0.0002, 0.00002\}. The batch size is selected from the set \{16, 32, 64, 128, 256, 512\}. Fig. \ref{Parameter Settings} shows the retrieval performance of the proposed method with varying learning rates and batch sizes in terms of mAP. For the three datasets, 0.0002 as the optimal learning rate for the best performance. Both too large and too small learning rates resulted in lower retrieval results. The best batch size was 128 for the three datasets. Adding more batch sizes that were larger than 128 caused a reduction in performance.

\subsection{Comparisons of the Related Methods}
\label{sec:Comparisons of the Related Methods}

In this section, we compare our retrieval result with state-of-the-art techniques in the computer vision field: deep fashion retrieval (DFR) \cite{liu2016deepfashion}, regional attention based deep feature (RADF) \cite{kim2018regional}, improved deep metric learning with multi-class n-pair loss objective (DML-MNP) \cite{sohn2016improved}, and deep metric learning with angular loss (DML-AL) \cite{wang2017deep}. All compared methods were based on ResNet-50, while Adam was used with a learning rate of 0.0002.

DFR is a model that aims to retrieve images of fashionable clothes. The loss function equals the cross-entropy loss plus triplet-margin loss with a weight balancing the two parts of the loss function. The triplet contains an anchor sample, a positive sample, and a negative sample. The 2,048 dimensional features extracted from ResNet-50 were then reduced into 512 dimensions by the FC layer. The Euclidean distance of features between the query image and retrieved images was computed to measure similarity. The RADF method used a context-aware regional attention network to address the problem of background clutter. The features extracted from RADF were 1,000 dimensions. We computed the Euclidean distance of features between the query image and retrieved images to evaluate the retrieval performance. DML-MNP used a new metric learning objective called multiclass n-pair loss, which generalizes triplet loss by allowing joint comparison among more than one negative example. This method addresses the problem of slow convergence that existing frameworks of DML based on contrastive loss and triplet loss often suffer from. DML-AL uses the angular loss for learning a better similarity metric, which takes the angle relationship into account. With the angular loss, scale invariance was introduced, and the robustness of the objective against feature variance improved. Features extracted from DML-MNP and DML-AL were both 128 dimensions, and the Euclidean distance of features between the query image and retrieved images was applied to similarity.

\subsubsection{Performance Changes with Small Training Samples}
\label{sec:Performance Changes with Small Training Samples}

\begin{figure}[htb]
\centering
{
\subfigure[UCMD] {\includegraphics[width=1.7 in,clip]{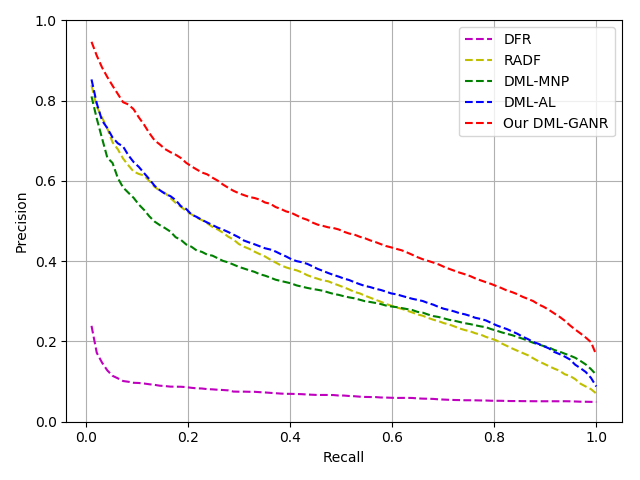}}
\subfigure[NWPU45] {\includegraphics[width=1.7 in,clip]{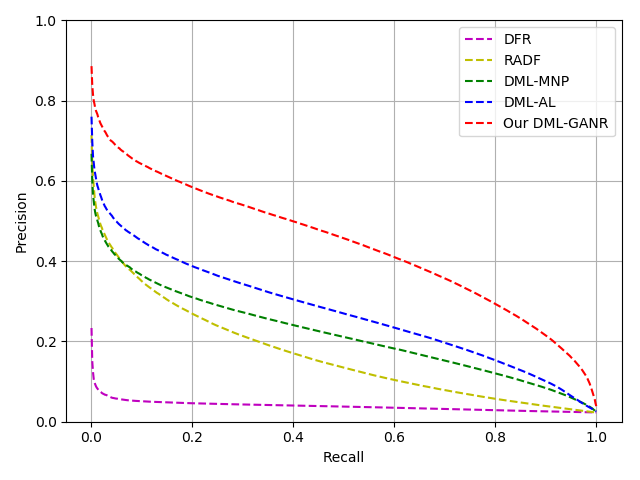}}
\subfigure[PatternNet] {\includegraphics[width=1.7 in,clip]{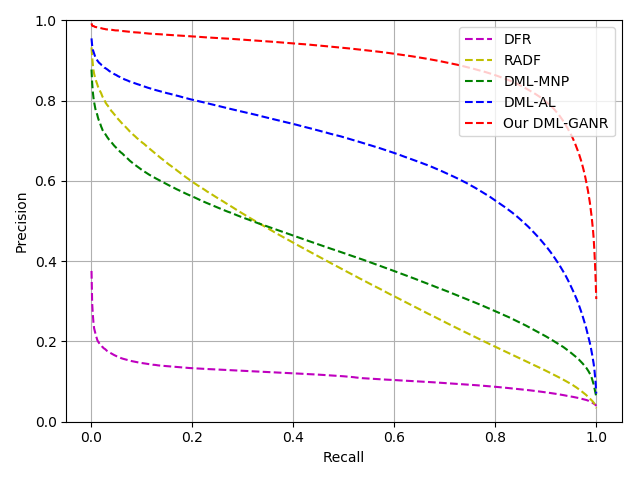}}
\subfigure[UCMD] {\includegraphics[width=1.7 in,clip]{./pics/UCMD_P2.png}}
\subfigure[NWPU45] {\includegraphics[width=1.7 in,clip]{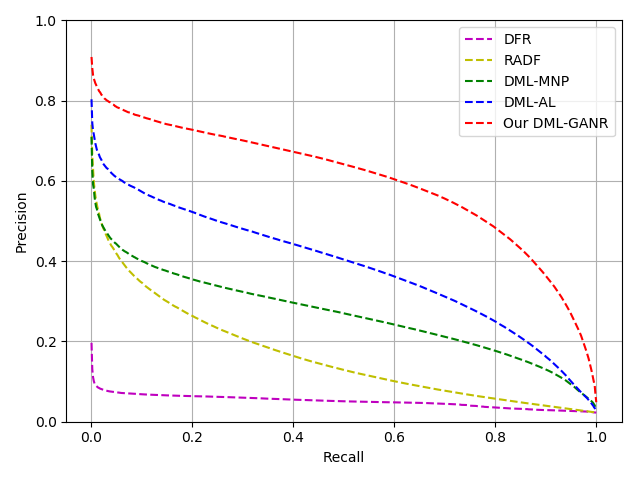}}
\subfigure[PatternNet] {\includegraphics[width=1.7 in,clip]{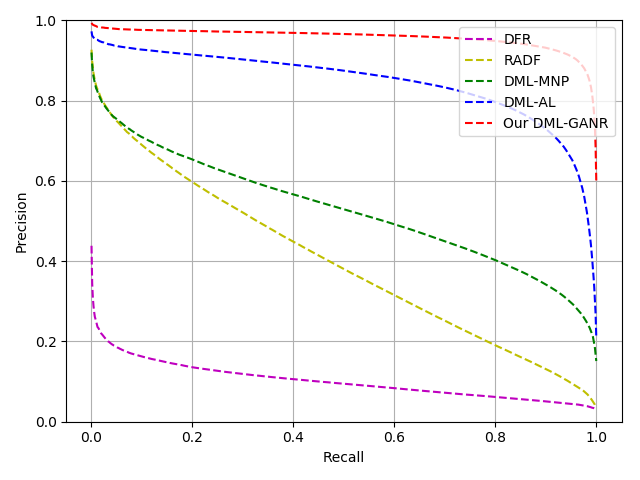}}
}
\caption{The precision-recall curves of different methods with small training samples on the three datasets. (a)-(c) 2\% (d)-(f) 5\%.}\label{PR_Small}
\end{figure}

To verify the advantages of DML-GANR on the small training samples, we randomly selected 2\% and 5\% of the images from each class of the three datasets. The corresponding 98\% and 95\% of the images were used for the query images. The image retrieval performances of 2\% and 5\% of the images were compared together with DFR, RADF, DML-MNP, and DML-AL. Table \ref{Number_small} shows the training number and testing number of the three datasets when 2\% and 5\% of the images from each class were selected as the training dataset. To clearly show the image retrieval performance variations with 2\% and 5\% of the images, we report the ANMRR, mAP, $\emph{P}@\emph{5}$, and $\emph{P}@\emph{50}$ in Table \ref{TSmallC}. The precision-recall curves of different methods with different percentages of the samples are shown in Fig. \ref{PR_Small}.

\renewcommand\arraystretch{1.25}
\begin{table}[htbp]
\caption{The Training Number and Testing Number of Three Datasets with Small Training Numbers.}
\centering{}%
\begin{tabular}{c|c |c|c |c|c |c}
\hline
Percentage & \multicolumn{2}{c|}{UCMD} & \multicolumn{2}{c|}{NWPU45} & \multicolumn{2}{c}{PatternNet}\\
\cline{2-7}
& training& test & training& test & training& test\\
\hline
2\%&	42&	2,058&	630&	30,870&	608	&29,792\\
5\%&	105&	1,995&	1,575&	29,925&	1,520&	28,880\\
\hline
\end{tabular}
\label{Number_small}
\end{table}

\renewcommand\arraystretch{1.25}
\begin{table*}[htbp]
\caption{The Results of Different Methods with Small Training Samples on UCMD, NWPU45, and PatternNet.}
\centering{}%
\begin{tabular}{c|c|c c c c| c c c c| c c c c}
\hline
Per	&Method& \multicolumn{4}{c|}{UCMD} & \multicolumn{4}{c|}{NWPU45} & \multicolumn{4}{c}{PatternNet}\\
\cline{3-14}
& & ANMRR & mAP & $\emph{P}@\emph{5}$  & $\emph{P}@\emph{50}$&ANMRR & mAP & $\emph{P}@\emph{5}$  & $\emph{P}@\emph{50}$&ANMRR & mAP & $\emph{P}@\emph{5}$  & $\emph{P}@\emph{50}$\\
\hline
2\%& DFR	&0.8102	&0.0772&	0.3026&	0.1152& 0.8804&	0.0401&	0.2829&	0.0913& 0.7624&	0.1087&	0.3878&	0.1982\\
&RADF	&0.4580&	0.3728&	0.8054	&0.5005& 0.6739	&0.1683	&0.6661	&0.4523 &0.4437	&0.3972	&0.8993	&0.7818\\
&DML-MNP &0.4874&	0.3431&	0.7432&	0.4440 &0.6258&	0.2039&	0.6225&	0.4279 &0.4328	&0.4186&	0.8469	&0.7115\\
&DML-AL	&0.4646&	0.3712&	0.7877&	0.4930 &0.5686&	0.2620	&0.6924&	0.5131& 0.2269&	0.6610&	0.9286	&0.8688\\
&DML-GANR	&\textbf{0.3342}&	\textbf{0.4900}&	\textbf{0.8803}&	\textbf{0.5971} & \textbf{0.3831}&	\textbf{0.4439}&	\textbf{0.8393}&	\textbf{0.6993}& \textbf{0.0560}	&\textbf{0.9001}	&\textbf{0.9881}&	\textbf{0.9746}\\
\hline
5\%	&DFR&	0.7942&	0.0956&	0.3293	&0.1348 &0.8778&	0.0487&	0.2754&	0.0990 &0.7594&	0.0996&	0.4183&	0.2268\\
&RADF	&0.4611&	0.3694	&0.8003&	0.4952 &0.6720&	0.1685&	0.6686	&0.4518& 0.4465&	0.3954&	0.8971	& 0.7769\\
&DML-MNP&	0.4173&	0.4103&	0.7609&	0.5061 &0.5560&	0.2587&	0.6393&	0.4611& 0.3199&	0.5204&	0.8837&	0.7685\\
&DML-AL&	0.4737&	0.3693&	0.7623&	0.4874 &0.4432	&0.3893&	0.7614	&0.6247 &0.1004&	0.8396&	0.9623	& 0.9352\\
&DML-GANR&	\textbf{0.2260}&	\textbf{0.6094}&	\textbf{0.9018}&	\textbf{0.6750} & \textbf{0.2616}&	\textbf{0.5887}&	\textbf{0.8778}&	\textbf{0.7854} & \textbf{0.0186}&	\textbf{0.9571}&	\textbf{0.9892}&	\textbf{0.9803}\\
\hline
\end{tabular}
\begin{tablenotes}
    \item $\emph{For ANMRR, lower values indicate better performance, while for mAP and P@k, larger is better}$.
\end{tablenotes}
\label{TSmallC}
\end{table*}

\begin{figure}[htb]
\centering
{
\subfigure[UCMD] {\includegraphics[width=2.3 in,clip]{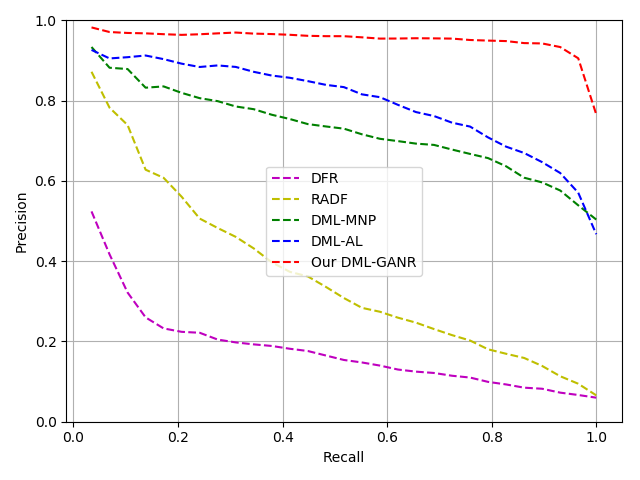}}
\subfigure[NWPU45] {\includegraphics[width=2.3 in,clip]{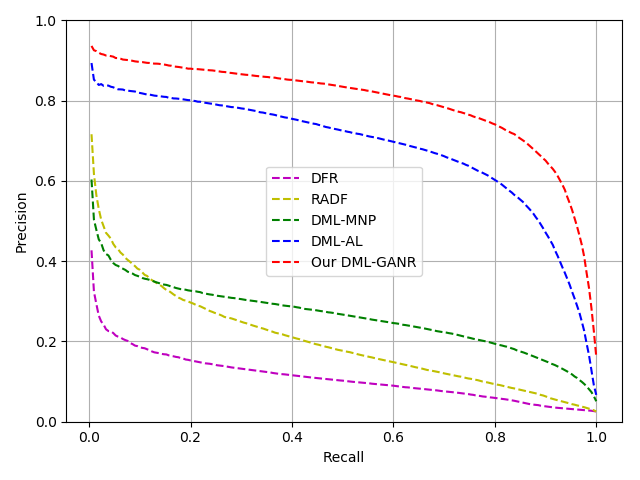}}
\subfigure[PatternNet] {\includegraphics[width=2.3 in,clip]{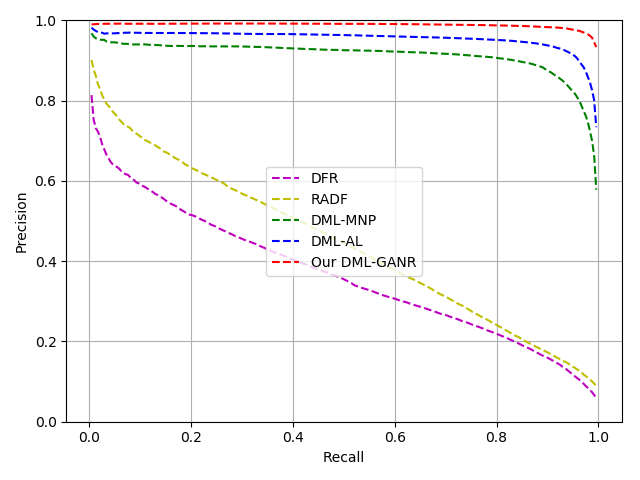}}
}
\caption{The precision-recall curves of different methods with 70\% of the samples on the three datasets.}\label{PR_other}
\end{figure}

As expected, the proposed DML-GANR method performed well on small training samples, as GAN can work in the case of small samples and mitigate the overfitting problem. The mAP, $\emph{P}@\emph{5}$, and $\emph{P}@\emph{50}$ are higher as performance is better. As shown in Table \ref{TSmallC}, our DML-GANR method was higher than other methods with small training samples in mAP, $\emph{P}@\emph{5}$, and $\emph{P}@\emph{50}$ for all three datasets. The ANMRR was lower as performance was better. DML-GANR was lower than other methods with small training samples in ANMRR. Especially for the PatternNet dataset, the mAPs of 2\% were 10.87\%, 39.72\%, 41.86\%, 66.10\%, and 90.01\%, corresponding to the retrieval results in DFR, RADF, DML-MNP, DML-AL, and DML-GANR respectively. DML-GANR significantly outperformed DFR by 79.14\%, which achieved the lowest value among compared methods. DML-GANR significantly outperformed DML-AL by 23.91\%, which achieved the highest value among compared methods. For the UCMD and NWPU45 dataset, $\emph{P}@\emph{5}$ of both 2\% and 5\% in our proposed DML-GANR were higher than 0.80. However, $\emph{P}@\emph{5}$ of both 2\% and 5\% of compared methods were lower than 0.80. For the PatternNet dataset, $\emph{P}@\emph{50}$ of both 2\% and 5\% in our proposed DML-GANR were higher than 0.95, while the highest value among compared methods was 0.9352. It was demonstrated that the compared methods suffer from overfitting problems. When meeting the small training samples, the performances of the compared methods worsened. Our proposed DML-GANR largely alleviated the overfitting problem and achieved good retrieval performance on the three datasets.

\subsubsection{Performance Changes with Other Training Samples}
\label{sec:Performance Changes with Other Training Samples}

To fully explore the performance of DML-GANR on large training samples compared with related methods, we randomly selected 10\%, 30\%, 50\%, and 70\% of the images from each class of the three datasets, and the corresponding 90\%, 70\%, 50\%, and 30\% of the images were used for the query images. The image retrieval performances of 10\%, 30\%, 50\%, and 70\% of the images were compared together with DFR, RADF, DML-MNP, and DML-AL. Table \ref{Number_other} shows the training number and testing number of the three datasets when 10\%, 30\%, 50\%, and 70\% of the images from each class were selected as the training dataset. To clearly show the image retrieval performance variations with 10\%, 30\%, 50\%, and 70\% of the samples, we report the ANMRR, mAP, $\emph{P}@\emph{5}$, and $\emph{P}@\emph{50}$ in Table \ref{TOtherC}. The precision-recall curves of different methods with 70\% of the samples are shown in Fig. \ref{PR_other}.

\renewcommand\arraystretch{1.25}
\begin{table}[htbp]
\caption{The Training Number and Testing Number of Three Datasets with Other Training Numbers.}
\centering{}%
\begin{tabular}{c|c |c|c |c|c |c}
\hline
Percentage & \multicolumn{2}{c|}{UCMD} & \multicolumn{2}{c|}{NWPU45} & \multicolumn{2}{c}{PatternNet}\\
\cline{2-7}
& training& test & training& test & training& test\\
\hline
10\%&	210&	1,890&	3,150&	28,350&	3,040&	27,360\\
30\%&	630	&1,470&	9,450&	22,050&	9,120&	21,280\\
50\%&	1,050&	1,050&	15,750&	15,750&	15,200&	15,200\\
70\%&	1,470&	630&	22,050&	9,450&	21,280&	9,120\\
\hline
\end{tabular}
\label{Number_other}
\end{table}

\renewcommand\arraystretch{1.25}
\begin{table*}[htbp]
\caption{The Results of Different Methods with Other Training Samples on UCMD, NWPU45, and PatternNet.}
\centering{}
\begin{tabular}{c|c|c c c c| c c c c| c c c c}
\hline
Per	&Method& \multicolumn{4}{c|}{UCMD} & \multicolumn{4}{c|}{NWPU45} & \multicolumn{4}{c}{PatternNet}\\
\cline{3-14}
& & ANMRR & mAP & $\emph{P}@\emph{5}$  & $\emph{P}@\emph{50}$&ANMRR & mAP & $\emph{P}@\emph{5}$  & $\emph{P}@\emph{50}$&ANMRR & mAP & $\emph{P}@\emph{5}$  & $\emph{P}@\emph{50}$\\
\hline
10\%& DFR&0.7790&	0.0915&	0.3468&	0.1271& 0.8686&	0.0473&	0.2779	&0.0975& 0.6646	&0.1711	&0.5155&	0.3353\\
&RADF&	0.4243&	0.4028&	0.8089&	0.5143 &0.6637&	0.1745&	0.6689&	0.4517 &0.4485&	0.3931&	0.8976&	0.7759\\
&DML-MNP	&0.4439&	0.3859&	0.7182&	0.4649& 0.6601&	0.1751&	0.5834&	0.3734 &0.2922&	0.5606&	0.9017&	0.7987\\
&DML-AL	&0.4453&	0.3935&	0.7873&	0.5130 &0.3475&	0.5112&	0.8101	&0.7109 &0.0618	&0.9016&	0.9722&	0.9552\\
&DML-GANR& \textbf{0.1574}&	\textbf{0.7080}	&\textbf{0.9294}&\textbf{0.7714} &\textbf{0.2004}&	\textbf{0.6773}	&\textbf{0.9009}	&\textbf{0.8306} &\textbf{0.0179}&	\textbf{0.9588}&	\textbf{0.9912}	&\textbf{0.9827}\\
\hline
30\%& DFR& 0.7687&	0.1170&	0.3525&	0.1544& 0.8378&	0.0584&	0.3214&	0.1373& 0.5284&	0.2784&	0.6873&	0.5059\\
&RADF&	0.4477&	0.3837&	0.7811&	0.4547& 0.6667&	0.1742&	0.6491&	0.4275& 0.4418&	0.4015	&0.8910&	0.7577\\
&DML-MNP&	0.3979&	0.4491&	0.7257&	0.4870& 0.6257&	0.2080	&0.5822	&0.3794& 0.3170&0.5267	&0.8581&	0.7314\\
&DML-AL&	0.2253&	0.6512&	0.8748&	0.7038& 0.2360&	0.6607&	0.8555&	0.7949& 0.0493&	0.9220	&0.9756&	0.9607\\
&DML-GANR&	\textbf{0.0541}&	\textbf{0.8863}&	\textbf{0.9631}&	\textbf{0.9020}& \textbf{0.1750}&	\textbf{0.7301}&	\textbf{0.9154}	&\textbf{0.8688}& \textbf{0.0101}	&\textbf{0.9755}&	\textbf{0.9940}&	\textbf{0.9895}\\
\hline
50\%& DFR&	0.7751&	0.0962&	0.3251&	0.1208& 0.7875&	0.0909&	0.3770&	0.1857& 0.5049&	0.3044&	0.7125&	0.5299\\
&RADF&	0.4489&	0.3851&	0.7619&	0.3921 &0.6693&	0.1702&	0.6425&	0.4046& 0.3852&	0.4612&	0.8792&	0.7236\\
&DML-MNP&	0.3587&	0.4688&	0.7333&	0.4737 &0.6147&	0.2207&	0.5687&	0.3668& 0.2994&	0.5580&	0.8857&	0.7627\\
&DML-AL&	0.1558&	0.7534&	0.9036&	0.7303& 0.2391&	0.6496&	0.8511&	0.7842& 0.0340&	0.9424&	0.9777&	0.9667\\
&DML-GANR&	\textbf{0.0361}&	\textbf{0.9183}&	\textbf{0.9726}&	\textbf{0.8874}&\textbf{0.1506}	&\textbf{0.7535}	&\textbf{0.9167}&	\textbf{0.8715}& \textbf{0.0093}&	\textbf{0.9793}&	\textbf{0.9947}&	\textbf{0.9907}\\
\hline
70\%& DFR&	0.6942&	0.1518&	0.3686&	0.1527& 0.7711&	0.1008&	0.3926	&0.1932 &0.4486&	0.3659&	0.7663&	0.5746\\
&RADF&	0.4801&	0.3639&	0.6873&	0.2783 &0.6093&	0.2199&	0.6589	&0.4168& 0.3932&	0.4518&	0.8861&	0.7151\\
&DML-MNP&	0.1974&	0.6596&	0.8102&	0.4528& 0.5547&	0.2739&	0.5720&	0.3894& 0.0329&	0.9261&	0.9697&	0.9521\\
&DML-AL&	0.1356&	0.7875&	0.9073&	0.5029& 0.2232&	0.6785&	0.8597&	0.7976 &0.0300&	0.9500&	0.9792	&0.9693\\
&DML-GANR&	\textbf{0.0284}&	\textbf{0.9343}&	\textbf{0.9727}&	\textbf{0.5765}& \textbf{0.1132}&	\textbf{0.8095}&	\textbf{0.9345}&	\textbf{0.8937} &\textbf{0.0042}&	\textbf{0.9885}&	\textbf{0.9962}&	\textbf{0.9936}\\
\hline
\end{tabular}
\begin{tablenotes}
    \item $\emph{For ANMRR, lower values indicate better performance, while for mAP and P@k, larger is better}$.
\end{tablenotes}
\label{TOtherC}
\end{table*}

As shown in Table \ref{TOtherC}, our DML-GANR method was higher than other methods with 10\%, 30\%, 50\%, and 70\% of the samples in mAP, $\emph{P}@\emph{5}$, and $\emph{P}@\emph{50}$ for all three datasets, where mAP, $\emph{P}@\emph{5}$, and $\emph{P}@\emph{50}$ were higher as performances were better. DML-GANR was lower than other methods with 10\%, 30\%, 50\%, and 70\% of the samples in ANMRR for all three datasets, where ANMRR was lower as performance was better. In Fig.\ref{PR_other}, the precision-recall curve of our DML-GANR method was superior to that of the compared methods in 70\% of the images. Overall, the proposed DML-GANR was accurate and efficient.
For the UCMD dataset, the mAP of 70\% of our DML-GANR was 93.43\%, while the highest value of the compared methods was only 78.75\%. The $\emph{P}@\emph{5}$ of 10\% of our DML-GANR was 0.9294, while the highest value of 70\% of the compared methods was only 0.7873. For the NWPU45 dataset, the mAP of 70\% of our DML-GANR was 80.95\%, while the highest value of 70\% of the compared methods was only 67.85\%. The $\emph{P}@\emph{5}$ values of our DML-GANR with 10\%, 30\%, 50\%, and 70\% of the  images were all higher than 0.90. However, the $\emph{P}@\emph{5}$ of DML-AL with 70\% of the images was only 0.8597, which as the highest $\emph{P}@\emph{5}$ value of all compared methods with 10\%, 30\%, 50\%, and 70\% of the images. For the PatternNet dataset, the mAPs of our DML-GANR with 10\%, 30\%, 50\%, and 70\% of the images were all higher than 95\%, and the $\emph{P}@\emph{5}$ of our DML-GANR with 10\%, 30\%, 50\%, and 70\% of the images were all higher than 0.99. However, the mAP of DML-AL with 70\% of the images as only 95\%, which was the highest mAP value of all compared methods with 10\%, 30\%, 50\%, and 70\% of the images. The $\emph{P}@\emph{5}$ of DML-AL with 70\% of the images as only 0.9792, which was the highest $\emph{P}@\emph{5}$ value of all compared methods with 10\%, 30\%, 50\%, and 70\% of the images.

\section{Conclusion}
\label{sec:Conclusion}
In this paper, we proposed a deep learning DML-GANR HSR-RSI retrieval method for a small number of labeled samples. Unlike previous retrieval methods, our proposed DML-GANR can perform well on the small training data samples, as GAN can work in the case of small samples and mitigate the overfitting problem. In DML-GANR, the features extracted from HFE are provided for DML and GAN, which constitute a novel and complete network. To make the features more representative, we fed the features extracted from each FC layer of the HFE into the DML. DML can minimize the intraclass variations and maximize the interclass variations, thereby, minimizing the distance between the image from the query domain and the images from the target domain. GAN can generate images that are similar to the real HSR-RSIs with the features extracted from the HFE as input. Therefore, the generated images can mitigate the overfitting problem and be a judge of whether the features extracted from the HFE are representative or not. Experiments in each part of DML-GANR were conducted. Therefore, the developed framework is reasonable and scalable for HSR-RSI retrieval, especially for the large HSR-RSI dataset. Compared to our method with the other methods both in small and other training samples, it was further demonstrated that our method has very competitive retrieval performance, especially on small training samples. Our method can contribute to classifying and recognizing ground objects in HSR-RSIs.

\renewcommand\refname{References}

\bibliographystyle{IEEEtran}
\bibliography{DML-GANR}
\end{document}